\documentclass[journal]{IEEEtran}
\usepackage{cite}

\usepackage{ifpdf}
 \ifpdf
 \else
 \fi

\ifCLASSINFOpdf
  \usepackage[pdftex]{graphicx}
  \graphicspath{{figures/}}
  \DeclareGraphicsExtensions{.pdf}
\else
  \usepackage[dvips]{graphicx}
  \graphicspath{{figures/}}
  \DeclareGraphicsExtensions{.eps}
\fi

\usepackage{amsmath}

\usepackage{algorithm}
\usepackage{algpseudocode}
\usepackage{array}

\ifCLASSOPTIONcompsoc
  \usepackage[caption=false, font=normalsize, labelfont=sf, textfont=sf]{subfig}
\else
  \usepackage[caption=false, font=footnotesize]{subfig}
\fi

\usepackage{url}

\usepackage{xcolor}
\usepackage{multirow}
\usepackage{bm}
\usepackage{hyperref}
\newtheorem{theorem}{Theorem}
\newtheorem{definition}{Definition}
\newtheorem{lemma}{Lemma}

\IEEEoverridecommandlockouts
\usepackage{tikz}
\usepackage{textcomp}
\usepackage{lipsum}

\newcommand\copyrighttext{%
	\footnotesize \textcopyright 2020 IEEE. Personal use of this material is permitted.  Permission from IEEE must be obtained for all other uses, in any current or future media, including reprinting/republishing this material for advertising or promotional purposes, creating new collective works, for resale or redistribution to servers or lists, or reuse of any copyrighted component of this work in other works. DOI: 10.1109/TCYB.2020.3000480}
\newcommand\copyrightnotice{%
	\begin{tikzpicture}[remember picture,overlay]
	\node[anchor=south,yshift=5pt] at (current page.south) {\fbox{\parbox{\dimexpr\textwidth-\fboxsep-\fboxrule\relax}{\copyrighttext}}};
	\end{tikzpicture}%
}

\begin{document}

\title{Toward a Controllable Disentanglement Network}

\author{Zengjie~Song,
        Oluwasanmi~Koyejo,
        and Jiangshe~Zhang
\thanks{Manuscript first version received July 3, 2019; resubmitted January 22, 2020; revised May 23, 2020. First version log number is CYB-E-2019-07-1382. Part of this work was done while Z. Song was a visiting Ph.D. student with the University of Illinois at Urbana-Champaign. This work is supported in part by the National Natural Science Foundation of China under Grant 61572393, Grant 11671317, Grant 61877049, and Grant 61976174, and in part by the China Scholarship Council. \textit{(Corresponding author: Jiangshe Zhang.)}}%
\thanks{Z. Song and J. Zhang are with the School of Mathematics and Statistics, Xi'an Jiaotong University, Xi'an 710049, China (e-mail: zjsong@hotmail.com; jszhang@mail.xjtu.edu.cn).}
\thanks{O. Koyejo is with the Department of Computer Science, University of Illinois at Urbana-Champaign, Urbana, IL 61801 USA (e-mail: sanmi@illinois.edu).}}

%
%

\markboth{Journal of \LaTeX\ Class Files,~Vol.~07, No.~03, July 2019}%
{Song \MakeLowercase{\textit{et al.}}: Toward a Controllable Disentanglement Network}
%

\maketitle
\copyrightnotice

\begin{abstract}
This paper addresses two crucial problems of learning disentangled image representations, namely controlling the degree of disentanglement during image editing, and balancing the disentanglement strength and the reconstruction quality. To encourage disentanglement, we devise a distance covariance based decorrelation regularization. Further, for the reconstruction step, our model leverages a soft target representation combined with the latent image code. By exploring the real-valued space of the soft target representation, we are able to synthesize novel images with the designated properties. To improve the perceptual quality of images generated by autoencoder (AE)-based models, we extend the encoder-decoder architecture with the generative adversarial network (GAN) by collapsing the AE decoder and the GAN generator into one. We also design a classification based protocol to quantitatively evaluate the disentanglement strength of our model. Experimental results showcase the benefits of the proposed model.
\end{abstract}


%
\IEEEpeerreviewmaketitle

\section{Introduction}\label{sec:intro}
\IEEEPARstart{O}{ne} of the long-standing challenges in machine learning community is to learn interpretable and robust representations of sensory data. Disentangling the hidden factors of variation provides the possibility of overcoming such a challenge \cite{Bengio13,Ridgeway16}. In a disentangled (or factorial) image representation, the generative factors of images correspond to independent subsets of the latent dimensions, such that changing a single factor causes a change in a single latent unit while being invariant to others \cite{Dupont18}. For example, a disentangled representation of face images could contain a set of latent units, each of which is sensitive to a specific facial attribute (i.e., generative factor) such as gender, age or wearing eyeglasses \cite{Lample17,Dupont18}. According to Lake \textit{et al.} \cite{Lake17}, disentangled representations have the potential to boost the performance of state-of-the-art machine learning approaches in several situations, including transfer learning and zero-shot learning. Besides, it is claimed that such representations are more robust against adversarial attacks \cite{Alemi17,Chen18}, and are also beneficial to design more robust multi-stage reinforcement learning agents \cite{Higgins17a}. Other scenarios where disentangled representations could play a role, such as novelty detection and information compression, can be found in the work of Ridgeway \cite{Ridgeway16}. 

There is substantial literature on learning disentangled image representations with deep neural networks. Most of these models feature an explicit or implicit regularization to induce the non-correlation between hidden representations, and then implement image editing by tweaking the representation components of interest accordingly. Example models are FadNet \cite{Lample17} that resorts to an adversarial-like training strategy \cite{Schmidhuber92}, or the ones based on the cross covariance regularization \cite{Cheung15,Kageback17}. Recently, several attempts have been made to investigate the disentanglement ability of the variational autoencoder (VAE) \cite{Kingma14a,Rezende14}. To penalize the correlation between different dimensions in the representation, VAE-based models, such as $\beta$-VAE \cite{Higgins17b}, DIP-VAE \cite{Kumar18}, JointVAE \cite{Dupont18}, and $\beta$-TCVAE \cite{Chen18}, emphasize the importance of minimizing the Kullback–Leibler divergence between the well-designed approximate posterior distribution and the disentangled prior distribution (e.g., an isotropic unit Gaussian) over the latent variable. However, all of the VAE's suffer from the nuisance blurring effects as observed in output images. By contrast, another set of works extending the generative adversarial network (GAN) \cite{Goodfellow14} is able to generate realistic-looking images when handing disentanglement-related tasks, including IcGAN \cite{Perarnau16} for face image editing, DR-GAN \cite{Tran17} for pose-invariant face recognition, StarGAN \cite{Choi18} for image-to-image translation, and Soft-Gated Warping-GAN \cite{Dong18} for pose-guided person image synthesis. The key point underlying these models is that the GAN generator learns to map the input image (or the representation of input) to the target image as accurately as possible, provide that the label or the domain information is given. In that case, the adversarial training process implicitly makes the inferred representation uncorrelated to the label \cite{Perarnau16,Tran17}, or makes the input image and the domain information interdependent for synthesizing target images \cite{Choi18,Dong18}.

While the models mentioned above have shown promise in specific disentanglement tasks, they pay limited attention to simultaneously tackling the following two problems, which are very important but challenging as learning disentangled representations, namely controlling degree of disentanglement, and preserving image quality.

\begin{figure*}[t]
	\centering
	\includegraphics[width=\textwidth]{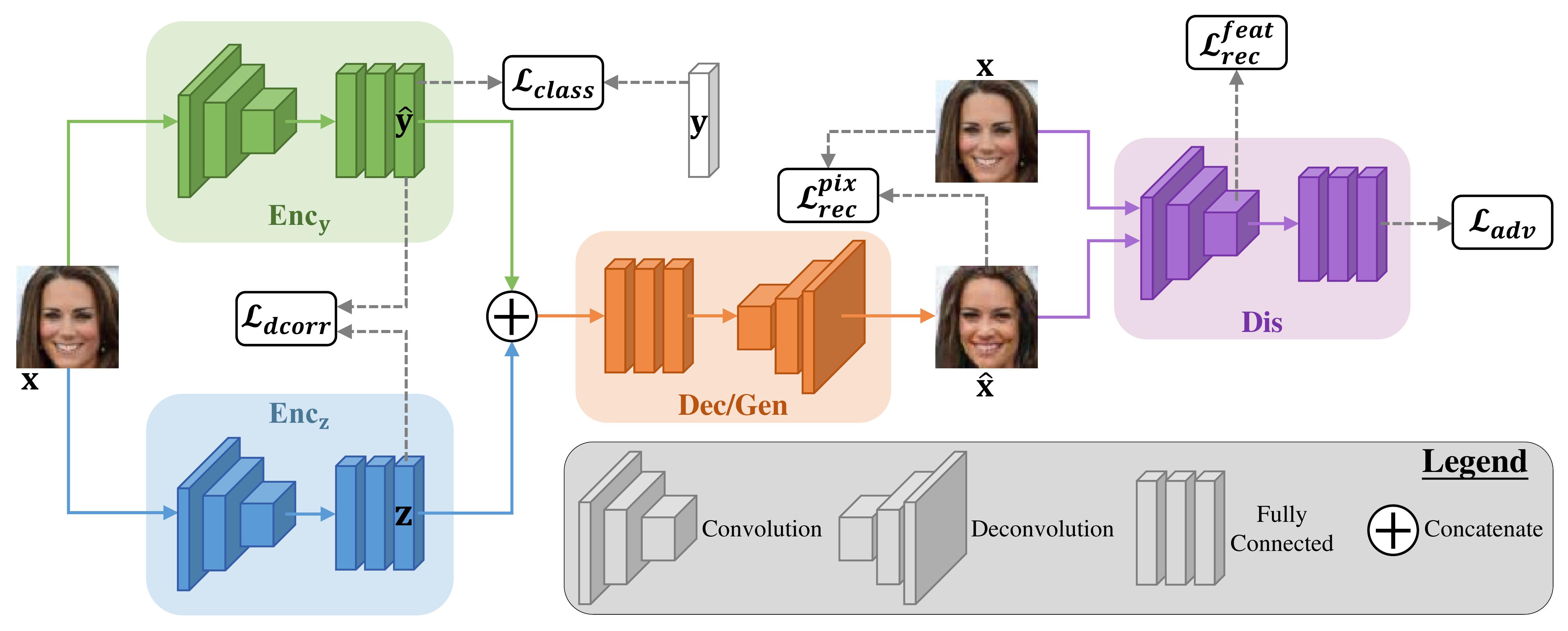}
	\caption{Network architecture of our CDNet for training. Solid lines denote the forward pass, and dashed lines indicate the flow to compute local losses. See Section \ref{sec:Model} for the detailed definitions of all symbols.}\label{fig:model_structure}
\end{figure*}

The first problem arises from controlling the degree of disentanglement during image editing, which means generating novel images with the designated attributes and, meanwhile, with the specific attribute intensities. For instance, given a face image, one may desire not only to synthesize a new smiling face, but also to synthesize a sequence of faces with expressions varying from no smile to toothy smile. Up to now, most of the existing disentanglement models \cite{Makhzani16,Larsen16,Perarnau16,Dumoulin17,Kim17,Donahue18,Engel18,Choi18}, however, mainly focus on whether the model can generate images with or without attributes of interest, rather than controlling the attribute intensities involved in the synthesized images. In practice, a more subtle manipulation of images is often most useful. This property could allow several potential applications, such as automatic face image editing and image color rendering \cite{Lample17}.

The second problem is how to generate images with the target attributes while preserving the core object identity and the image quality. Generally speaking, if the learned target representation part is not as independent as possible of other parts, changing one generative factor could induce changes of other factors, even falsifying the object identity. This phenomenon has been observed in many existing works \cite{Cheung15,Kulkarni15,Chen16,Higgins17b,Ma18,Kumar18,Chen18}. For instance, as shown in \cite{Chen18}, heightening the baldness intensity of a face concurrently leads to the visually-perceptible change of the object identity. Additionally, many disentanglement models based on autoencoder (AE) \cite{Cheung15,Kulkarni15,Higgins17b,Kumar18,Dupont18,Chen18} usually produce blurry output images, thus giving rise to the image quality degradation.

In this paper, we present a simple yet effective model named as Controllable Disentanglement Network, or CDNet, to address the aforementioned two problems. The overall network architecture of our model for training is depicted in Fig. \ref{fig:model_structure}. Specifically, CDNet combines the AE with the GAN to construct a new deep neural network, which is further divided into four parts: $\text{Enc}_\mathbf{y}$, $\text{Enc}_\mathbf{z}$, $\text{Dec}/\text{Gen}$, and $\text{Dis}$. The two encoders $\text{Enc}_\mathbf{y}$ and $\text{Enc}_\mathbf{z}$ are used to learn the soft target representation $\mathbf{\hat{y}}$ and the latent representation $\mathbf{z}$, respectively. The representation $\mathbf{\hat{y}}$ acts to capture class- or attribute-related information by solving the supervised classification subtask, while the representation $\mathbf{z}$ serves to extract information different from those in $\mathbf{\hat{y}}$ via the proposed decorrelation regularization. The decoder of the AE (labeled $\text{Dec}$) works to reconstruct input image $\mathbf{x}$ when given the two groups of representations $\mathbf{\hat{y}}$ and $\mathbf{z}$. By viewing the AE decoder as the GAN generator (labeled $\text{Gen}$), the reconstructed image $\mathbf{\hat{x}}$ is also treated as the fake image, with the purpose of fooling the discriminator (labeled $\text{Dis}$) such that the $\text{Dis}$ cannot distinguish the fake image from the real image. To achieve image editing, only the two well-trained encoders and the decoder are utilized, and modifications of $\mathbf{\hat{y}}$ values are performed with regard to the designated classes or attributes.

In summary, we highlight our contributions as follows.
\begin{enumerate}
	\item To encourage disentanglement, a novel decorrelation regularization based on the distance covariance \cite{Szekely07} is proposed to learn two independent representations, i.e., the soft target representation $\mathbf{\hat{y}}$ and the latent representation $\mathbf{z}$ (see Section \ref{sec:decorr_loss} for details).
	\item To implement the disentanglement controlling, we explore the potential of the soft target representation, rather than the discrete label as used in many existing works \cite{Mirza14,Cheung15,Perarnau16,Makhzani16}, to reconstruct original and synthesize new images. This soft target representation is a probability representation at training time, and its element scale implicitly indicates how much class or attribute information is included in input image. Under this setting, one is able to decrease or increase the specific element scales to modify attribute intensities of the synthesized image at testing phase (see Section \ref{sec:method_manipulate} for details).
	\item To improve the perceptual quality of the generated images, we extend the AE architecture with GAN, where the GAN generator and the AE decoder are tied as the same one by parameter sharing and joint training. This model combination is inspired by the VAE/GAN \cite{Larsen16}, with the difference in that we build the CDNet based on the deterministic AE, and introduce a parallel encoder to learn the soft target representation. The new integrated model shows improved ability to reconstruct images and learn disentangled representations (see Sections \ref{sec:recons_ability}, \ref{sec:disentangle_ability}, and \ref{sec:disentangle_degree} for empirical comparisons).
	\item To quantitatively compare the disentanglement strength of our model, an evaluation protocol is designed. To our best knowledge, this is the first work that leverages classification to analyze how the effect of representation scales with disentanglement performance (see Section \ref{sec:disentangle_class} for details).
\end{enumerate}

The PyTorch source code of our model is available at \href{https://github.com/zjsong/CDNet}{https://github.com/zjsong/CDNet}.

\section{Related Work}\label{sec:related_work}
Due to its nature of interpretability and robustness, disentangled representations have been attracting increasing attention in recent years. Here, we divide the related disentanglement models into three prominent groups: AE-based models \cite{Cheung15,Kulkarni15,Higgins17b,Hou17,Lample17,Kumar18}, GAN-based models \cite{Chen16,Kim17,Donahue18,Ma18,Choi18}, and integrations of AE's and GAN's \cite{Makhzani16,Larsen16,Perarnau16,Dumoulin17,Engel18}. In the following, we provide a detailed survey of this topic from these three perspectives.

\subsection{AE-based Models} \ By constraining the latent variables to be invariant to image attributes, the basic AE model can be extended to handle the disentanglement task. For instance, the FadNet \cite{Lample17} learns a classifier to predict the attribute given the latent representation, while the latent representation inferred by the encoder tries to prevent the classifier from predicting the correct attribute values. This adversarial-like process enables the model, in an implicit manner, to learn a latent representation containing information different from the attributes. By contrast, Cheung \textit{et al.} \cite{Cheung15} employ an explicit decorrelation regularization, based on the cross covariance (XCov), to approach the same goal. Our CDNet is also built on the basic AE, but with three notable differences from the aforementioned models. First, in contrast to the AE-based models that take advantage of the class or attribute label during image editing, the CDNet implements image editing by using the inferred soft target representation, thus our model is applicable to scenarios where label information is unavailable. Second, compared with the XCov regularization, the distance covariance (dCov) we use for disentanglement encourages statistical independence rather than non-correlation between variables \cite{Szekely07}, leading to a stronger disentanglement ability (see Sections \ref{sec:disentangle_class} and \ref{sec:ablation_study} for details). Third, the usage of a GAN in our model also markedly improves the perceptual quality of the output images.

The vanilla VAE \cite{Kingma14a,Rezende14} has been shown to learn disentangled representations, but with limited disentanglement ability on simple datasets such as FreyFaces or MNIST. Higgins \textit{et al.} \cite{Higgins17b} and Kumar \textit{et al.} \cite{Kumar18} refine VAE to learn controllable disentangled factors, implemented by putting implicit independence constraints on the approximate posterior over latent variables. Kulkarni \textit{et al.} \cite{Kulkarni15} achieve disentanglement based on a special training scheme, where pairs of rendered images that differ only in one factor of variation are provided. Another VAE-like model \cite{Hou17} utilizes the vector arithmetic technique \cite{Mikolov13} to control attribute intensities. We note that those models can be trained stably in general, however, they are prone to obtain blurry images. Besides, the unsupervised learning strategy adopted in many of these models cannot guarantee the non-correlation between learned representations, and thus the change of one attribute (e.g., smiling) may induce changes of other attributes (e.g., hairstyle or azimuth) as observed in \cite{Kumar18}. In fact, Locatello \textit{et al.} \cite{Locatello19} have theoretically shown that the unsupervised disentanglement learning is fundamentally impossible without inductive biases both on models and datasets. This conclusion indicates that the role of supervision is crucial \cite{Locatello19}, which is coincident with the idea of using labeled training data in our model.

\subsection{GAN-based Models} \ The plain GAN \cite{Goodfellow14} does not show any apparent disentanglement properties, nevertheless, subsequent works have enhanced GANs. Donahue \textit{et al.} \cite{Donahue18} propose the semantically decomposed GANs that learn to decompose the latent code into an identity-related portion and observation-related portion, thus modifying face images by varying the observation vector. Focusing on person image generation, Ma \textit{et al.} \cite{Ma18} use an adversarial network to learn mappings from Gaussian noise to the embedding feature space, which provides more control over the foreground, background, and pose information of the input image. In the InfoGAN \cite{Chen16}, a subset of facial attributes is changed by manipulating the learned categorical codes, but with no conspicuous visual difference among generated images (such as the ``Hair style'' variation shown therein). By coupling two GANs together, the DiscoGAN \cite{Kim17} leverages the cross-domain relations to perform the facial attribute conversion task. The StarGAN \cite{Choi18}, one of the state-of-the-art multi-domain image translation models, achieves facial attribute manipulation by a single generator and shows remarkable ability to synthesize high quality images. However, StarGAN's such superiority is obtained only when the number of operated attribute domains is small, and thus it is less effective for balancing the disentanglement strength and the reconstruction quality across a mass of different attributes (e.g., all 40 facial attributes in CelebA face images, as illustrated in the supplementary material). Moreover, it's worth noting that many GAN-based models still suffer from the problems of training instability and model collapse \cite{Salimans16}, which also make the disentangled representation learning more challenging.

\subsection{Combined AE and GAN Models} \ A natural way to alleviate aforementioned problems is to combine AE with GAN, thereby leveraging both models' strengths in a complementary manner. To approach this goal, several existing works explore the adversarial training strategy in the latent space of AEs. The main point of these models is making the GAN discriminator indistinguishable 1) between the aggregated posterior of the latent variable and an arbitrary prior \cite{Makhzani16}; or 2) between samples in latent space and encoded data (rather than prior samples) \cite{Engel18}; or 3) between joint samples of the data and the corresponding latent variable from the encoder and joint samples from the decoder \cite{Dumoulin17}.

The IcGAN \cite{Perarnau16} and the VAE/GAN \cite{Larsen16} are another two works falling into this line of literature. In IcGAN, an encoder is added into the conditional GAN \cite{Mirza14} to learn a mapping from the image space to the representation space, and thus implementing image editing by changing the conditional information inferred from the real image. In VAE/GAN, the VAE decoder and the GAN generator are viewed as the same mapping by parameter sharing and joint training, and the GAN discriminator acts to measure sample similarity in the feature space. Although the network architecture is similar, the representation learning method of our CDNet differs substantially from these models. First, the attribute representations learned by IcGAN and VAE/GAN are still correlated with each other, degrading the ability to manipulate images subtly. By contrast, the decorrelation regularization facilitates the CDNet to learn independent representations, which enables the model to control disentanglement at testing phase. Second, both of the pixel reconstruction error and the feature reconstruction error are explored in CDNet, and thus the stability of model training and the perceptual quality of output images are all improved. We conducted a series of experiments to compare these two models with the proposed CDNet in Sections \ref{sec:recons_ability}, \ref{sec:disentangle_ability}, and \ref{sec:disentangle_degree}.

\section{Methodology}\label{sec:Model}
We propose the CDNet, a novel model combining AE with GAN, that advances the state-of-the-art toward jointly solving the problems of the image disentanglement controlling and the image quality balance between disentanglement and reconstruction. The CDNet architecture is shown in Fig. \ref{fig:model_structure}, and it consists of four components: $\text{Enc}_\mathbf{y}$, $\text{Enc}_\mathbf{z}$, $\text{Dec}/\text{Gen}$, and $\text{Dis}$. Specifically, the encoder $\text{Enc}_\mathbf{y}$ aims to learn the \textit{soft target representation} $\mathbf{\hat{y}}$ to extract class or attribute information from the discrete label $\mathbf{y}$. This goal is approached by training $\text{Enc}_\mathbf{y}$ to solve a supervised classification task. Another encoder $\text{Enc}_\mathbf{z}$ serves to learn the \textit{latent representation} $\mathbf{z}$ under two constraints: being informative to reconstruct input image $\mathbf{x}$ and being uncorrelated with (even independent of) $\mathbf{\hat{y}}$. Here the independence between $\mathbf{\hat{y}}$ and $\mathbf{z}$ is induced by the proposed decorrelation regularization. The $\text{Dec}$ takes as input the representations $\mathbf{\hat{y}}$ and $\mathbf{z}$ to reconstruct input image. Because CDNet collapses the AE decoder and the GAN generator into one, the reconstructed image $\mathbf{\hat{x}}$ is also treated as the fake image generated by the $\text{Gen}$. With this setting, we train the $\text{Dis}$ to distinguish fake image $\mathbf{\hat{x}}$ from real image $\mathbf{x}$, and also use the middle layer representations of the $\text{Dis}$ to compute the feature reconstruction error.

In the following, we first formulate all local losses used to train different network components. Then we derive the integrated loss function and elaborate the associated training algorithm of our CDNet. After that, several practical considerations for implementation are provided. Finally, the method to manipulate images with controllable disentanglement is illustrated in two applications.

In addition to the aforementioned symbols, let $\mathbf{X}$ denote the mini-batch version of the input image $\mathbf{x}$, $N$ the mini-batch size, and similarly define $\mathbf{\hat{X}}$, $\mathbf{Y}$, $\mathbf{\hat{Y}}$, and $\mathbf{Z}$ for the reconstructed/fake image $\mathbf{\hat{x}}$, label $\mathbf{y}$, soft target representation $\mathbf{\hat{y}}$, and latent representation $\mathbf{z}$, respectively.

\subsection{Four Local Losses}\label{sec:local_losses}
\subsubsection{Classification Loss $\mathcal{L}_{class}$}\label{sec:class_loss}
Minimizing the classification loss $\mathcal{L}_{class}$ enables the encoder $\text{Enc}_\mathbf{y}$ to inject the class or attribute information into the soft target representation $\mathbf{\hat{y}}$. We select two classification loss functions to fit the following two application cases, respectively.

\textit{Case 1:} For the multiclass scenario (e.g., a handwritten digit belongs to only one of the 10 classes), we first use the softmax nonlinearity to scale each element of $\mathbf{\hat{y}}$. Then the cross entropy between the discrete labels $\mathbf{Y}$ and the soft target representations $\mathbf{\hat{Y}}$ is computed as the classification loss.

\textit{Case 2:} For the multilabel scenario (e.g., face images with or without smiling, eyeglasses, and blond hair attributes), we first use the sigmoid nonlinearity to scale each element of $\mathbf{\hat{y}}$. Then the binary cross entropy between $\mathbf{Y}$ and $\mathbf{\hat{Y}}$ is derived as the classification loss.

\subsubsection{Decorrelation Loss $\mathcal{L}_{dcorr}$}\label{sec:decorr_loss}
We propose to leverage the distance covariance (dCov) \cite{Szekely07} based regularization to learn the latent representation $\mathbf{z}$, which is expected to be independent of the soft target representation $\mathbf{\hat{y}}$. To obtain this decorrelation loss (or regularization), we first compute the $N$ by $N$ distance matrices $(a_{n,m})$ and $(b_{n,m})$ containing all pairwise distances:
\begin{align*}
a_{n,m} &= \| \mathbf{\hat{y}}_{n} - \mathbf{\hat{y}}_{m} \|_{2}, \quad n,m = 1, 2, \dots, N,\\
b_{n,m} &= \| \mathbf{z}_{n} - \mathbf{z}_{m} \|_{2}, \quad n,m = 1, 2, \dots, N
\end{align*}
where $\|\cdot\|_{2}$ is the $l^{2}$-norm. Then take all doubly centered distances:
\begin{align*}
A_{n,m} &:= a_{n,m} - \bar{a}_{n\cdot} - \bar{a}_{\cdot m} + \bar{a}_{\cdot\cdot},\\
B_{n,m} &:= b_{n,m} - \bar{b}_{n\cdot} - \bar{b}_{\cdot m} + \bar{b}_{\cdot\cdot}
\end{align*}
where $\bar{a}_{n\cdot}$ is the $n$th row mean, $\bar{a}_{\cdot m}$ is the $m$th column mean, and $\bar{a}_{\cdot\cdot}$ is the grand mean of the distance matrix $(a_{n,m})$. The notation is similar for the $b$ values. Finally, the squared sample distance covariance, treated as our decorrelation loss, is simply the arithmetic average of the products $A_{n,m}B_{n,m}$:
\begin{equation}\label{eqn:loss_dcorr_dCov}
\mathcal{L}_{dcorr} = \text{dCov}^{2}(\mathbf{\hat{Y}}, \mathbf{Z}) = \frac{1}{N^{2}}\sum_{n=1}^{N}\sum_{m=1}^{N}A_{n,m}B_{n,m}.
\end{equation}

By minimizing the decorrelation loss in Eq. \eqref{eqn:loss_dcorr_dCov} to approach zero, the soft target representation $\mathbf{\hat{y}}$ and the latent representation $\mathbf{z}$ would tend to be independent. This conclusion is supported by the following Theorem \ref{thm:independ_dcov}. Let $\mathbf{U}$ and $\mathbf{V}$ denote two random vectors, $\text{dCov}^{2}_{K}(\mathbf{U}, \mathbf{V})$ is the squared sample distance covariance between $\mathbf{U}$ and $\mathbf{V}$, and $K$ indicates the number of pairwise sample points\footnote{Both $\text{dCov}^{2}$ and $\text{dCov}^{2}_{K}$ stand for the squared sample distance covariance, and the subscript $K$ is used to facilitate theoretical analysis.}.
\begin{theorem}\label{thm:independ_dcov}
Suppose two random vectors $\mathbf{U}$ and $\mathbf{V}$ satisfy $E(\|\mathbf{U}\|_{2}) < \infty$ and $E(\|\mathbf{V}\|_{2}) < \infty$. If $\lim_{K \to \infty} \text{dCov}^{2}_{K}(\mathbf{U}, \mathbf{V}) = 0$, then almost surely $\mathbf{U}$ and $\mathbf{V}$ are independent.
\end{theorem}
\begin{IEEEproof}
The proof can be established by using the Definition 3, Theorem 2, and Theorem 3 in \cite{Szekely07}. See the Appendix for completeness.
\end{IEEEproof}

By comparison, Cheung \textit{et al.} \cite{Cheung15} use the following cross covariance (XCov) to facilitate disentanglement:
\begin{equation}\label{eqn:loss_dcorr_XCov}
\text{XCov}(\mathbf{\hat{Y}}, \mathbf{Z}) = \frac{1}{2}\sum_{i,j}\left[\frac{1}{N}\sum_{n=1}^{N}(\hat{y}_{n,i}-\bar{\hat{y}}_{\cdot i})(z_{n,j}-\bar{z}_{\cdot j})\right]^{2}
\end{equation}
where $\bar{\hat{y}}_{\cdot i}=\frac{1}{N}\sum_{n=1}^{N}\hat{y}_{n,i}$ and $\bar{z}_{\cdot j}=\frac{1}{N}\sum_{n=1}^{N}z_{n,j}$. It is worth emphasizing that one of the most important differences between $\text{dCov}^{2}$ and XCov is that, minimizing $\text{dCov}^{2}$ encourages the \textit{independence} between two random variables, while minimizing XCov encourages the \textit{non-correlation}. In this regard, the $\text{dCov}^{2}$ should induce stronger disentanglement than the XCov. Additionally, our model is also compatible with XCov, and replacing $\text{dCov}^{2}$ with XCov in CDNet also achieves improved disentanglement performance over the model in \cite{Cheung15} (see Sections \ref{sec:disentangle_ability} and \ref{sec:disentangle_degree} for empirical comparisons).

\subsubsection{Reconstruction Loss $\mathcal{L}_{rec}$}\label{sec:recons_loss}
The reconstruction principle of CDNet is similar to the basic AE, that is, the soft target representations $\mathbf{\hat{Y}}$ and the latent representations $\mathbf{Z}$ are first computed by the two encoders, respectively, and then fed to the decoder to reconstruct original images:
\begin{equation}
\mathbf{\hat{Y}} = \text{Enc}_\mathbf{y}(\mathbf{X}), \ \ \mathbf{Z} = \text{Enc}_\mathbf{z}(\mathbf{X}), \ \ \mathbf{\hat{X}} = \text{Dec}(\mathbf{\hat{Y}}, \mathbf{Z}).
\end{equation}
We use the reconstruction loss to measure the difference between original images $\mathbf{X}$ and reconstructions $\mathbf{\hat{X}}$. A common choice is the mean squared error (MSE) computed in pixel space:
\begin{equation}\label{eqn:loss_rec_pixel}
\mathcal{L}_{rec}^{pix} = \frac{1}{NM}\sum_{n=1}^{N}\|\mathbf{x}_{n} - \mathbf{\hat{x}}_{n}\|_{2}^{2}
\end{equation}
where $M$ indicates the image dimensionality.

As discussed in Section \ref{sec:related_work}, AE's trained with the pixel-level MSE usually produce blurring effects in the resulting image, which is also observed in Section \ref{sec:recons_ability} in this paper. We conjecture that lacking the meaningful spatial correlation properties of original images causes the quality degradation of reconstructions. The hidden representation of a deep convolutional neural network, however, can extract such spatial correlation properties from the input image \cite{Larsen16,Hou17}. Inspired by this observation, we compute the feature-matching difference \cite{Larsen16} as the additional reconstruction loss:
\begin{equation}\label{eqn:loss_rec_feature}
\mathcal{L}_{rec}^{feat} = \frac{1}{ND_{l}}\sum_{n=1}^{N}\|h_{l}(\mathbf{x}_{n}) - h_{l}(\mathbf{\hat{x}}_{n})\|_{2}^{2}
\end{equation}
where $D_{l}$ is the dimensionality of the $l$th layer of the GAN discriminator and $h_{l}(\mathbf{x})$ denotes the hidden representation of $\mathbf{x}$ at that layer. It's worth noting that the perceptual loss \cite{Gatys16,Johnson16}, computed by some pretrained high-performing CNN such as VGG \cite{Simonyan15}, can also measure the high-level feature difference between two images. But the perceptual loss is not widely applicable when the training data are not in image format (e.g., audio or text data), or when the training data are images but the size of image is smaller than the acceptable image size to the pretrained model. By comparison, the feature-matching difference loss \eqref{eqn:loss_rec_feature} is customized and learned on the fly at each iteration step, and thus enabling our model to be flexible and extensible to handle data disentanglement tasks.

The final reconstruction loss consists of two parts:
\begin{equation}\label{eqn:loss_rec}
\mathcal{L}_{rec} = \mathcal{L}_{rec}^{pix} + \lambda_{rec}\mathcal{L}_{rec}^{feat}
\end{equation}
where $\lambda_{rec}$ controls the trade-off between reconstructions of global features (i.e., $\mathcal{L}_{rec}^{pix}$) and local details (i.e., $\mathcal{L}_{rec}^{feat}$). By minimizing these two local reconstruction losses together, the CDNet is enforced to restore identity-preserving images with high-level structures. In practice, we have observed that using the pixel-wise reconstruction loss also makes the adversarial training more stable.

\subsubsection{Adversarial Loss $\mathcal{L}_{adv}$}
The goal of incorporating GAN into CDNet is to improve the perceptual quality of the output images. In the GAN part of CDNet, the generator $\text{Gen}$ maps the inferred soft target representation and the latent representation to image space, while the discriminator $\text{Dis}$ estimates the probability that a sample belongs to the data distribution. The GAN is trained such that the $\text{Dis}$ can tell apart real from fake images, and meanwhile the $\text{Gen}$ can generate images that ``fool'' the $\text{Dis}$. To this end, we need to maximize/minimize the following adversarial loss
\begin{equation}\label{eqn:adv_loss}
\mathcal{L}_{adv} = \log(\text{Dis}(\mathbf{X})) + \log(1 - \text{Dis}(\text{Gen}(\mathbf{\hat{Y}}, \mathbf{Z})))
\end{equation}
with respect to the $\text{Dis}$/$\text{Gen}$.

Note that the exact choice of the GAN model (and so the adversarial loss) is not fundamental in our CDNet, since the plain GAN described here is adequate to improve the image perceptual quality. To obtain images with better visual fidelity, several advanced GAN models such as PatchGAN \cite{Li16} and WGAN-GP \cite{Gulrajani17} could be employed.

\subsection{Integrated Loss and Training Algorithm}
We train our combined model with the integrated loss
\begin{equation}\label{eqn:loss_total}
\mathcal{L} = \mathcal{L}_{class} + \lambda_{dcorr}\mathcal{L}_{dcorr} + \mathcal{L}_{rec} + \mathcal{L}_{adv}
\end{equation}
where the exact expression of each local loss is presented above, and $\lambda_{dcorr}$ balances the quality of the reconstruction and the strength of the disentanglement. All of the local losses included in Eq. \eqref{eqn:loss_total} are complementary to each other, enabling the CDNet to address the two crucial disentanglement problems mentioned in Section \ref{sec:intro}.

We train each model component of CDNet with the associated local losses. More specifically, we first train the encoder $\text{Enc}_\mathbf{y}$ to solve a supervised classification problem, which is implemented by minimizing the classification loss $\mathcal{L}_{class}$ w.r.t. $\bm{\theta}_{\text{Enc}_\mathbf{y}}$. After that, we fix the $\text{Enc}_\mathbf{y}$ and just use it to infer the soft target representation of input. For another encoder $\text{Enc}_\mathbf{z}$, the decorrelation loss $\mathcal{L}_{dcorr}$ and the reconstruction loss $\mathcal{L}_{rec}$ are minimized to update its parameters $\bm{\theta}_{\text{Enc}_\mathbf{z}}$. The parameters of the decoder (and so, the generator), $\bm{\theta}_{\text{Dec}}$, are modified based on the minimization of $\mathcal{L}_{rec}$ and $\mathcal{L}_{adv}$. The adversarial loss $\mathcal{L}_{adv}$ is also related to the discriminator, and is used to learn parameters $\bm{\theta}_{\text{Dis}}$. For clarity, we summarize the training algorithm for CDNet in Algorithm \ref{alg:training_CDNet}. More details about the parameter setting can be found in Section \ref{sec:train_detail}.
\begin{algorithm}[t]
	\linespread{1.1}\selectfont
	\caption{Training the CDNet}\label{alg:training_CDNet}
	\begin{algorithmic}[1]
		\State $\bm{\theta}_{\text{Enc}_\mathbf{y}}$, $\bm{\theta}_{\text{Enc}_\mathbf{z}}$, $\bm{\theta}_{\text{Dec}}$, $\bm{\theta}_{\text{Dis}} \gets$ initialize network parameters
		\Repeat \Comment{Train $\text{Enc}_\mathbf{y}$ independently}
		\State $\{\mathbf{X}, \mathbf{Y}\} \gets$ random mini-batch from dataset
		\State $\mathbf{\hat{Y}} \gets$ $\text{Enc}_\mathbf{y}(\mathbf{X})$
		\State $\mathcal{L}_{class} \gets$ (Binary)CrossEntropy($\mathbf{\hat{Y}}, \mathbf{Y}$)
		\State $\bm{\theta}_{\text{Enc}_\mathbf{y}} \stackrel{+}\leftarrow -\nabla_{\bm{\theta}_{\text{Enc}_\mathbf{y}}}\mathcal{L}_{class}$
		\Until{deadline}
		\Repeat \Comment{Train $\text{Enc}_\mathbf{z}$, $\text{Dec}$, and $\text{Dis}$ simultaneously}
		\State $\mathbf{X} \gets$ random mini-batch from dataset
		\State $\mathbf{\hat{Y}} \gets \text{Enc}_\mathbf{y}(\mathbf{X})$
		\State $\mathbf{Z} \gets \text{Enc}_\mathbf{z}(\mathbf{X})$
		\State $\mathcal{L}_{dcorr} \gets$ $\text{dCov}^{2}(\mathbf{\hat{Y}}, \mathbf{Z})$ or $\text{XCov}(\mathbf{\hat{Y}}, \mathbf{Z})$
		\State $\mathbf{\hat{X}} \gets \text{Dec}(\mathbf{\hat{Y}}, \mathbf{Z})$
		\State $\mathcal{L}^{pix}_{rec} \gets$ MSE($\mathbf{\hat{X}}, \mathbf{X}$)
		\State $\mathcal{L}^{feat}_{rec} \gets$ MSE($h_{l}(\mathbf{\hat{X}}), h_{l}(\mathbf{X})$)
		\State $\mathcal{L}_{rec} \gets \mathcal{L}_{rec}^{pix} + \lambda_{rec}\mathcal{L}_{rec}^{feat}$
		\State $\mathcal{L}_{adv} \gets \log(\text{Dis}(\mathbf{X})) + \log(1 - \text{Dis}(\mathbf{\hat{X}}))$
		\State $\bm{\theta}_{\text{Enc}_\mathbf{z}} \stackrel{+}\leftarrow -\nabla_{\bm{\theta}_{\text{Enc}_\mathbf{z}}}(\lambda_{dcorr}\mathcal{L}_{dcorr} + \mathcal{L}_{rec})$
		\State $\bm{\theta}_{\text{Dec}} \stackrel{+}\leftarrow -\nabla_{\bm{\theta}_{\text{Dec}}}(\mathcal{L}_{rec} + \lambda_{adv}\mathcal{L}_{adv})$
		\State $\bm{\theta}_{\text{Dis}} \stackrel{+}\leftarrow \nabla_{\bm{\theta}_{\text{Dis}}}\mathcal{L}_{adv}$
		\Until{deadline}
	\end{algorithmic}
\end{algorithm}

\subsection{Implementation Details}
In addition to the aforementioned recipe for training the CDNet, we also adopt the following techniques demonstrated in \cite{Lample17} to stabilize the training process in practice.
\begin{enumerate}
	\item \textit{Appending the soft target representation to each layer of the decoder:} The soft target representation inferred by the encoder $\text{Enc}_\mathbf{y}$ contains class- or attribute-related information. This discriminative information is utilized by the decoder in two different ways. For each of the fully-connected layers of the decoder, we concatenate the soft target representation and the hidden layer representation together as a whole input to the next layer. For all the convolutions of the decoder, we append the soft target representation as additional constant input channels.
	\item \textit{Decorrelation loss scheduling:} To avoid that the decorrelation regularization dominates the parameter updating of the encoder $\text{Enc}_\mathbf{z}$, which would destroy the model's reconstruction ability, we use a variable weight for the regularization parameter $\lambda_{dcorr}$. That is, we linearly increase the $\lambda_{dcorr}$ to a target value over the early training process, and then clamp it for the remaining training process. By doing so, the effect of decorrelation regularization is gradually imposed on the learning of the latent representation.
	\item \textit{Dropout:} We use the dropout \cite{Srivastava14} in all fully-connected layers, except the final layer, of the two encoders and the discriminator. In our experiments, we found that dropout is beneficial to prevent the encoder $\text{Enc}_\mathbf{y}$ and the discriminator $\text{Dis}$ from overfitting, and is also helpful for the encoder $\text{Enc}_\mathbf{z}$ to learn a latent representation being as independent as possible of the soft target representation.
\end{enumerate}

\subsection{Methods to Manipulate Images}\label{sec:method_manipulate}
For image editing, the key operation is to modify the value of the soft target representation $\mathbf{\hat{y}}$ accordingly. Based on the two application cases described in Section \ref{sec:class_loss}, we give two corresponding methods to manipulate images.

\begin{figure*}[t]
	\centering
	\includegraphics[width=\textwidth]{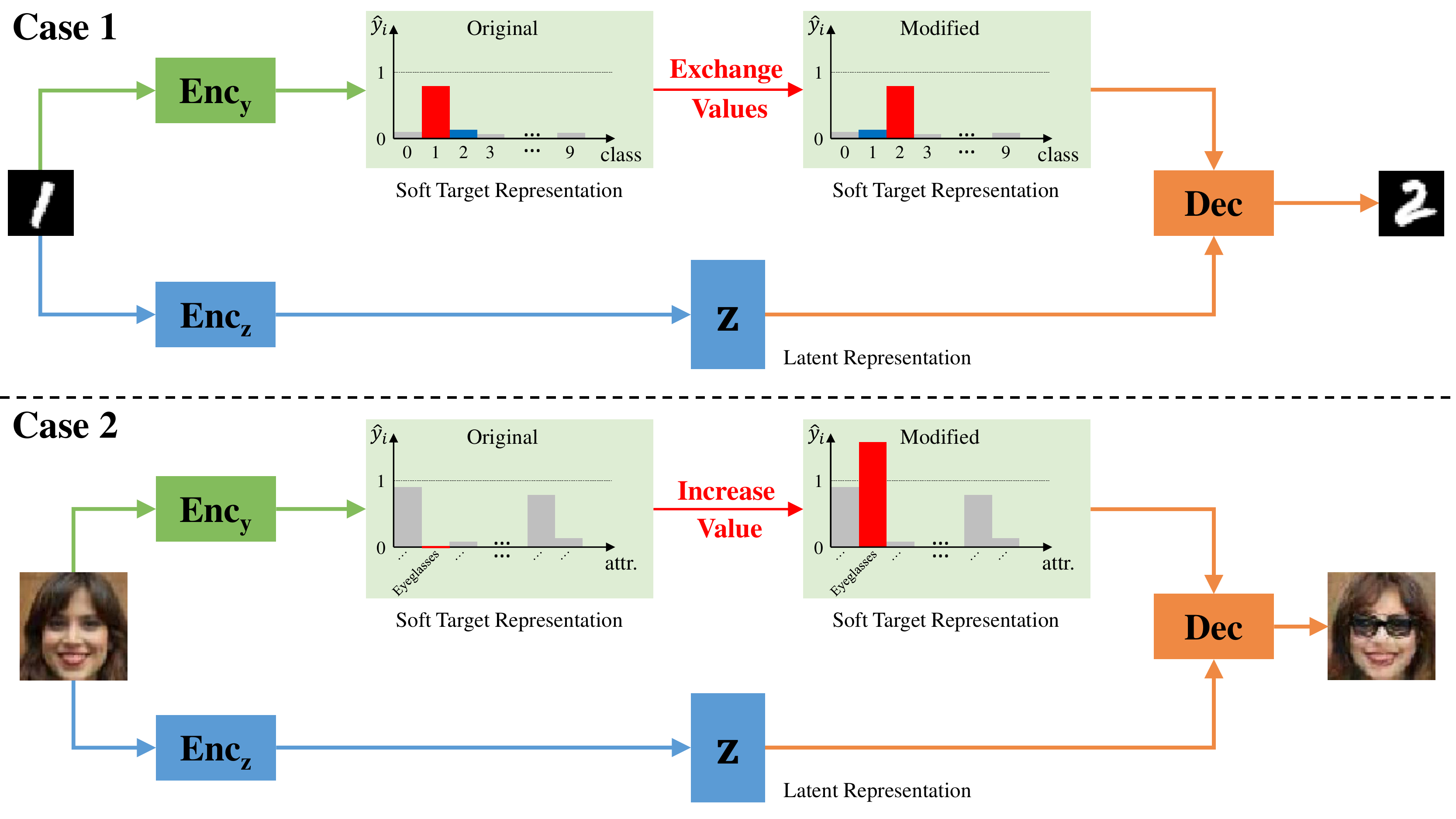}
	\caption{Manipulating images. \textit{Case 1} addresses the multiclass scenario where the mutual exclusion exists among multiple classes. \textit{Case 2} is for the multilabel scenario where multiple attributes are independent of each other. The detailed procedure is illustrated in Section \ref{sec:method_manipulate}. Best viewed in color.}\label{fig:manipulate_images}
\end{figure*}

In \textit{Case 1}, taking the handwritten digit as an example, we aim to generate a new digit with the handwriting style similar to the given digit. As shown in Fig. \ref{fig:manipulate_images}, we first employ the two encoders, $\text{Enc}_\mathbf{y}$ and $\text{Enc}_\mathbf{z}$, to infer the soft target representation $\mathbf{\hat{y}}$ and the latent representation $\mathbf{z}$ of the given digit ``1'' in boldface. Then we modify $\mathbf{\hat{y}}$ by exchanging the third element (corresponding to the digit 2 class) and the maximum element (ideally corresponding to the digit 1 class), while keeping remaining elements fixed. In this way, only two elements of $\mathbf{\hat{y}}$ at most are exchanged, and thus the representation structure with component summation of 1 is preserved. Finally, we feed the modified $\mathbf{\hat{y}}$ and the unchanged $\mathbf{z}$ to the decoder to generate the new digit ``2'' which is also in boldface.

In \textit{Case 2}, with the face image as an example, the goal is to synthesize a new face with the desired attribute and intensity while preserving the core identity. As we can see from Fig. \ref{fig:manipulate_images}, the overall procedure is similar to the first case, only with the difference in modifying $\mathbf{\hat{y}}$. Actually, in order to generate a new face with eyeglasses, we just replace the original (near) zero value corresponding to ``Eyeglasses'' attribute with the new value (e.g., 3.5) in $\mathbf{\hat{y}}$. Note that \textit{during image editing} the modified attribute value is not necessarily restricted in $[0, 1]$, meaning it can also take other real values outside this interval\footnote{We empirically found that the extended interval $[-2, 5]$ is large enough for our model to generate new face images with various facial attribute intensities.}. Specifically, the small value (near 0 or less than 0) indicates that the synthesized image tends to exclude the target attribute, while the big value (near 1 or greater than 1) implies that the synthesized image prefers containing that attribute. In this regard, the continuous attribute value can be viewed as a sliding knob, the magnitude of which controls how much a specific attribute could be perceivable in the final image. As illustrated in \cite{Chen16}, by modifying attribute values in such an exaggerated way, the soft target representation is able to cover a wide range that the network was never trained on and we will get meaningful generalization (see Section \ref{sec:disentangle_degree} for empirical evidence).

\section{Experiments}
We conduct five groups of experiments to substantiate the benefits of our CDNet model. First, we show that the combination of AE and GAN is able to improve the quality of reconstructed images. Second, we verify that the proposed decorrelation regularization and the image manipulation methods are competent to disentangle factors of variation. Third, by synthesizing images with various attributes and attribute intensities, we qualitatively illustrate the CDNet's ability to control the degree of disentanglement. Fourthly, we propose a classification based protocol to quantitatively compare the disentanglement strength of the CDNet. Finally, we perform an ablation study to investigate the effectiveness of different loss terms. Additional results are provided in the supplementary material. Before presenting our experimental results, we introduce the experimental setup in detail.

\subsection{Experimental Setup}
\subsubsection{Datasets}
We use two representative datasets for the two application cases described in Section \ref{sec:class_loss}. The first one is MNIST \cite{LeCun98}, which contains 70,000 grayscale handwritten digit images with $28\times28$ pixels for each and scaled to $[0, 1]$. We randomly split the dataset into 50,000 training, 10,000 validation, and 10,000 test samples, respectively. The discrete label has the one-hot vector form. The second dataset is CelebA \cite{Liu15}, which consists of 202,599 RGB face images of celebrities. For pre-processing, all face images are first center-cropped and then downsampled to $64\times64$ RGB pixels and scaled to $[-1, 1]$. We use $80\%$ images for training, $10\%$ for validation, and $10\%$ for test as used in several earlier works. Additionally, the discrete label is represented by a binary vector with dimensionality 40, where each dimension corresponds to one attribute with value 1 indicating containing that attribute and 0 not.

\subsubsection{Baselines}
We evaluate our CDNet with three baselines which have similar network architectures to ours, and we give all models as follows.
\begin{enumerate}
	\item AE-XCov \cite{Cheung15}: a pure AE-based model that leverages the cross covariance (XCov) to learn the middle-layer uncorrelated representations.
	\item IcGAN \cite{Perarnau16}: a model introducing an encoder to GAN to implement the inference mechanism, and it achieves disentanglement via modifying the discrete labels inferred by the encoder.
	\item VAE/GAN \cite{Larsen16}: a model that combines a VAE with a GAN, and its reconstruction error is derived only in the high-level feature space. The disentangled representation is computed using the vector arithmetic method in \cite{Mikolov13}.
	\item CDNet-XCov (ours): an instantiation of our CDNet model, where the XCov is used as the decorelation loss.
	\item CDNet-dCov (ours): another instantiation of the CDNet model, which utilizes the distance covariance (dCov)-based regularization, proposed in this paper, to learn independent representations.
\end{enumerate}
The AE-XCov serves to illustrate the problem of blurring effects produced by the pure AE-based models. The IcGAN and VAE/GAN are selected to confirm that the decorrelation regularization employed in our model is beneficial to learn disentangled representations, and that the integration of both the pixel-wise and the feature-wise reconstruction errors is helpful to improve the reconstruction quality. The CDNet-dCov is compared with the CDNet-XCov to empirically verify that under the same settings, the dCov regularization can facilitate models to generate easier-perceptible attributes than the XCov.  

\subsubsection{Training Details}\label{sec:train_detail}
We implement the IcGAN and the VAE/GAN based on the network architectures as shown in \cite{Perarnau16} and \cite{Larsen16}, respectively. For the CDNet, the two encoders have the same architecture, which consists of convolution layers followed by fully-connected layers. The main part of decoder is symmetric to encoder, but using deconvolution (a.k.a. transposed convolution or fractional striding) \cite{Radford16} for the up-sampling. The AE-XCov is built with the similar way to CDNet, except that we append the soft target representation to each layer of the CDNet decoder to strength the influence of class or attribute information on output images. The architecture details can be found in the supplementary material.

We perform the hyperparameter selection according to the validation-set performance. Specifically, for both of the two datasets, $\lambda_{rec}$ takes the value of 1 to balance the two reconstruction errors. The $\lambda_{adv}$ appeared in Algorithm \ref{alg:training_CDNet} weights the reconstruction ability of decoder v.s. fooling the discriminator, equaling to 0.01. For the MNIST dataset, $\lambda_{dcorr}$ is linearly increased to 1 over the first 50,000 iterations, while for the CelebA dataset, $\lambda_{dcorr}$ is gradually increased until it reaches 0.05 across the first 50,000 iterations. The CDNet models are trained with the RMSProp optimizer, where we set the learning rate $=0.0001$ and a batch size of 100 for MNIST, the learning rate $=0.00005$ and a batch size of 128 for CelebA. With a single NVIDIA GeForce GTX 1080 GPU, training CDNet-XCov takes about 1.33 hours on MNIST and 11.17 hours on CelebA; for CDNet-dCov, the training time is about 1.34 hours on MNIST and 11.23 hours on CelebA.

\begin{figure}[!t]
	\centering
	\subfloat[]{\includegraphics[width=0.45\textwidth]{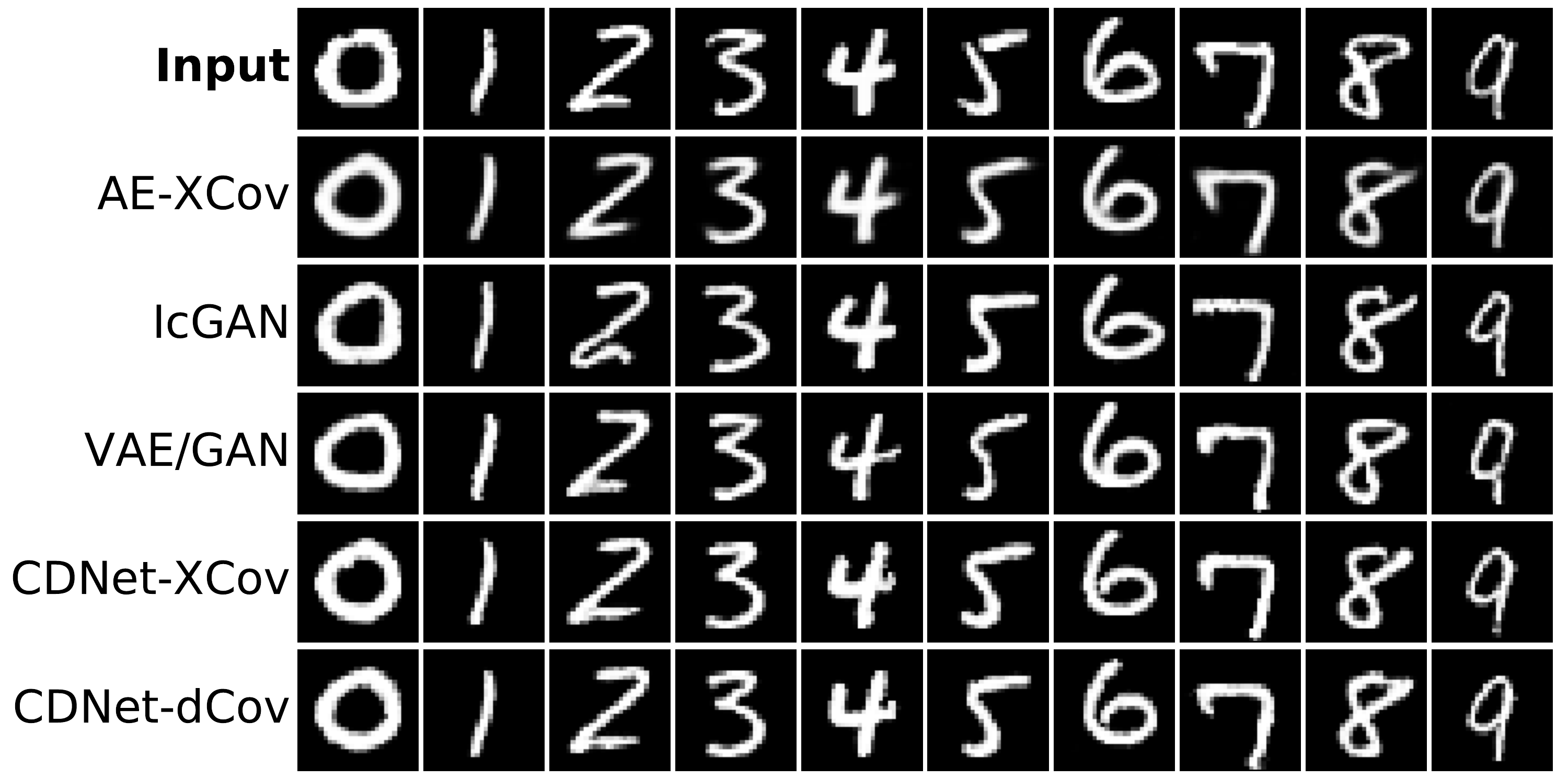}}\\ \vspace{-2mm}
	\subfloat[]{\includegraphics[width=0.45\textwidth]{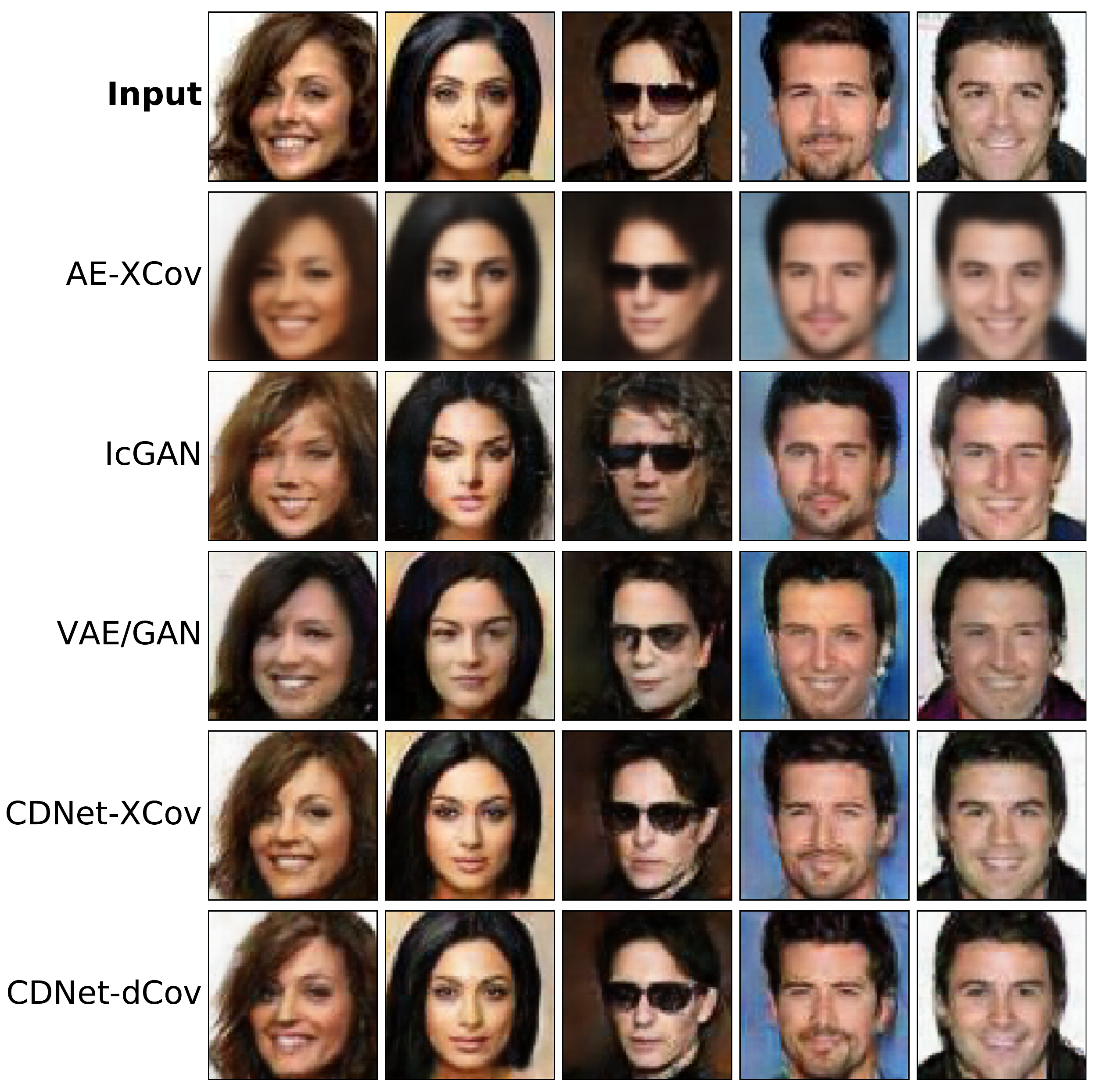}}%
	\caption{Reconstructions of (a) handwritten digit images from MNIST test set and (b) face images from CelebA test set. Best viewed in color.}\label{fig:recons_imgs}
\end{figure}
\subsection{Reconstruction}\label{sec:recons_ability}
We first visualize the reconstruction results to make a qualitative comparison between different models. As shown in Fig. \ref{fig:recons_imgs} (a), when reconstructing handwritten digit images from MNIST, all models consistently perform well in terms of the visual perception quality. However, the performance gap becomes large as reconstructing face images from CelebA. As we can see from Fig. \ref{fig:recons_imgs} (b), reconstructions of the AE-XCov contain the main structure of the input face image, but lose the details due to the blurring effect. The IcGAN on the contrary works well to recover texture features, while apparently falsifying the core object identity in reconstructed images. Although the VAE/GAN approaches a balance between the reconstruction accuracy and the visual fidelity, it cannot recover some specific local features, such as hair texture and eyeglasses. By contrast, our two CDNet models can reconstruct images with higher visual fidelity. The results demonstrate that as a combination of AE and GAN, the CDNet enjoys two advantages for reconstruction tasks, i.e., preserving the core object identity and simultaneously recovering local detail features.

To quantitatively evaluate the reconstruction ability of the proposed model, we utilize three well-known image quality assessment indexes, namely root-mean-square error (RMSE), peak signal-to-noise ratio (PSNR), and multi-scale structure similarity (SSIM) \cite{Wang03}, to assess the reconstructed images' quality. As can be seen from Table \ref{tab:recons_quality}, the two instantiations of the CDNet significantly outperform the other models across all three evaluation metrics. We conclude that in the CDNet, the integration of the pixel-wise reconstruction error (from AE) and the feature-wise reconstruction error (from GAN) provides an effective way to improve the reconstruction quality.
\begin{table}[t]
	\renewcommand{\arraystretch}{1.3}
	\centering
	\caption{Reconstruction Quality on CelebA Test Set. The Results Are Formatted as Mean ($\pm$Standard Deviation). Best Two Scores in Each Column Are Highlighted in Bold}\label{tab:recons_quality}
	\begin{tabular}{|l||*{3}{c|}}
		\hline
		Model               		 &RMSE           		  &PSNR           		   &SSIM   \\
		\hline\hline
		\multirow{2}{*}{AE-XCov}     &0.0991                  &20.2640                 &0.8701          \\
									 &($\pm$0.0209)           &($\pm$1.7993)           &($\pm$0.0585)   \\
		\hline
		\multirow{2}{*}{IcGAN}       &0.1668                  &15.7877                 &0.7236          \\
									 &($\pm$0.0388)  		  &($\pm$1.9966)           &($\pm$0.1247)   \\
		\hline
		\multirow{2}{*}{VAE/GAN}     &0.1403          		  &17.2311                 &0.7804          \\
									 &($\pm$0.0287)  		  &($\pm$1.7287)           &($\pm$0.0921)   \\
		\hline
		\multirow{2}{*}{CDNet-XCov}  &\textbf{0.0834}         &\textbf{21.7564}        &\textbf{0.9099} \\
									 &($\pm$0.0175)  		  &($\pm$1.7683)  		   &($\pm$0.0377)   \\
		\hline
		\multirow{2}{*}{CDNet-dCov}  &\textbf{0.0828}         &\textbf{21.8181}        &\textbf{0.9108} \\
									 &($\pm$0.0172)  		  &($\pm$1.7580) 		   &($\pm$0.0377)   \\
		\hline
	\end{tabular}
\end{table}

\subsection{Disentanglement}\label{sec:disentangle_ability}
\begin{figure*}[!t]
	\centering
	\subfloat[]{\includegraphics[width=0.19\textwidth]{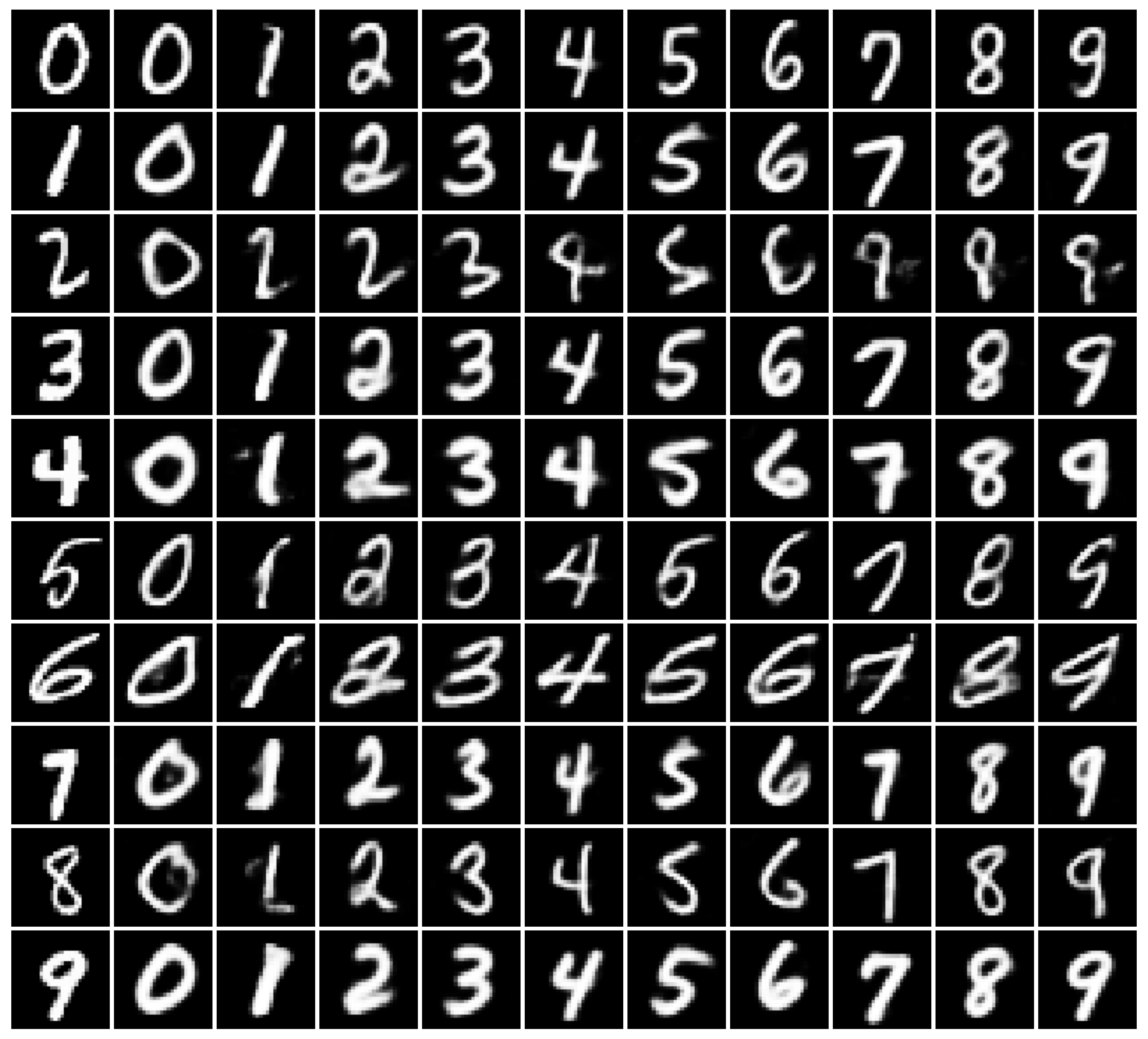}}\hspace{0.1cm}
	\subfloat[]{\includegraphics[width=0.19\textwidth]{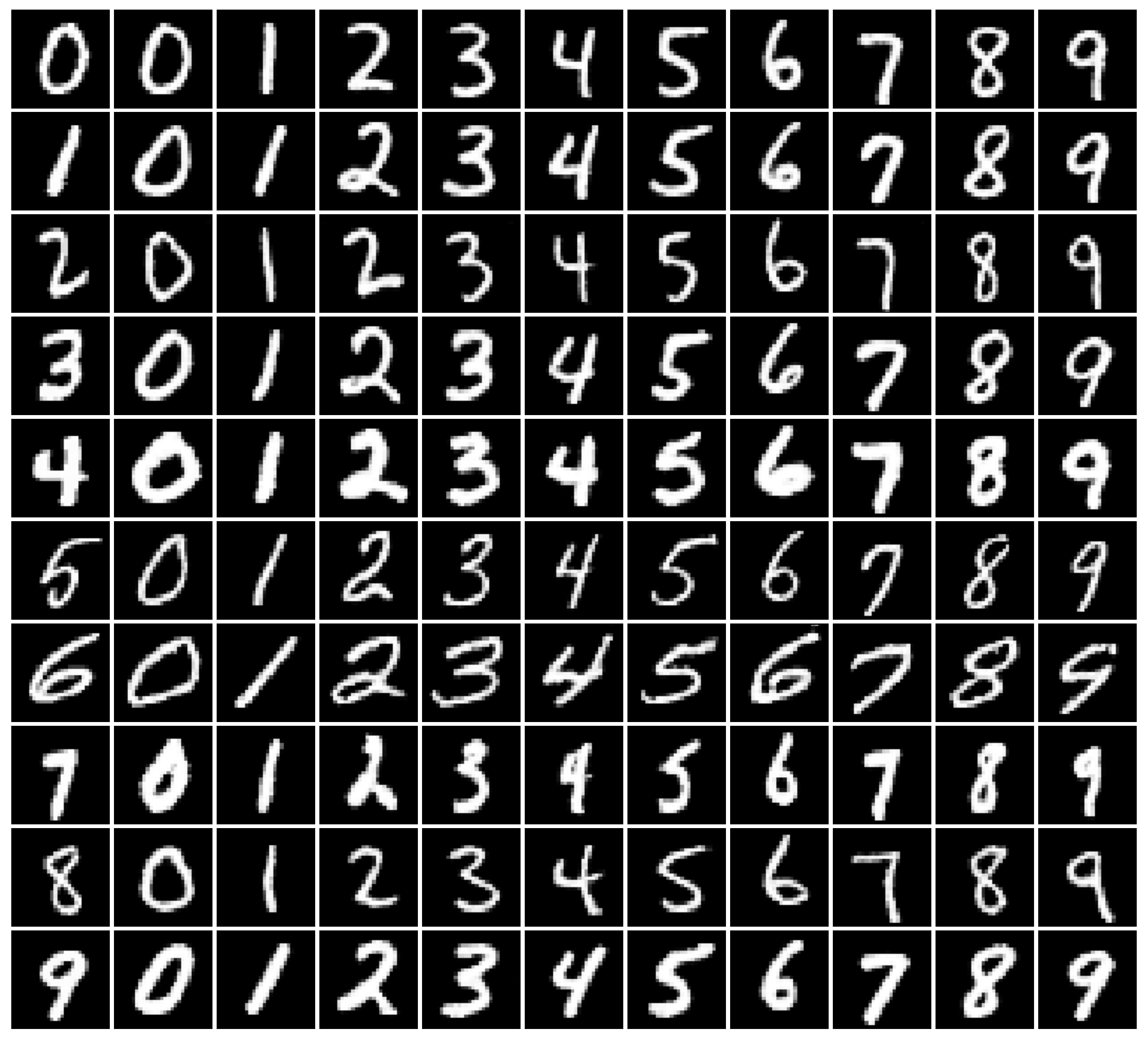}}\hspace{0.1cm}
	\subfloat[]{\includegraphics[width=0.19\textwidth]{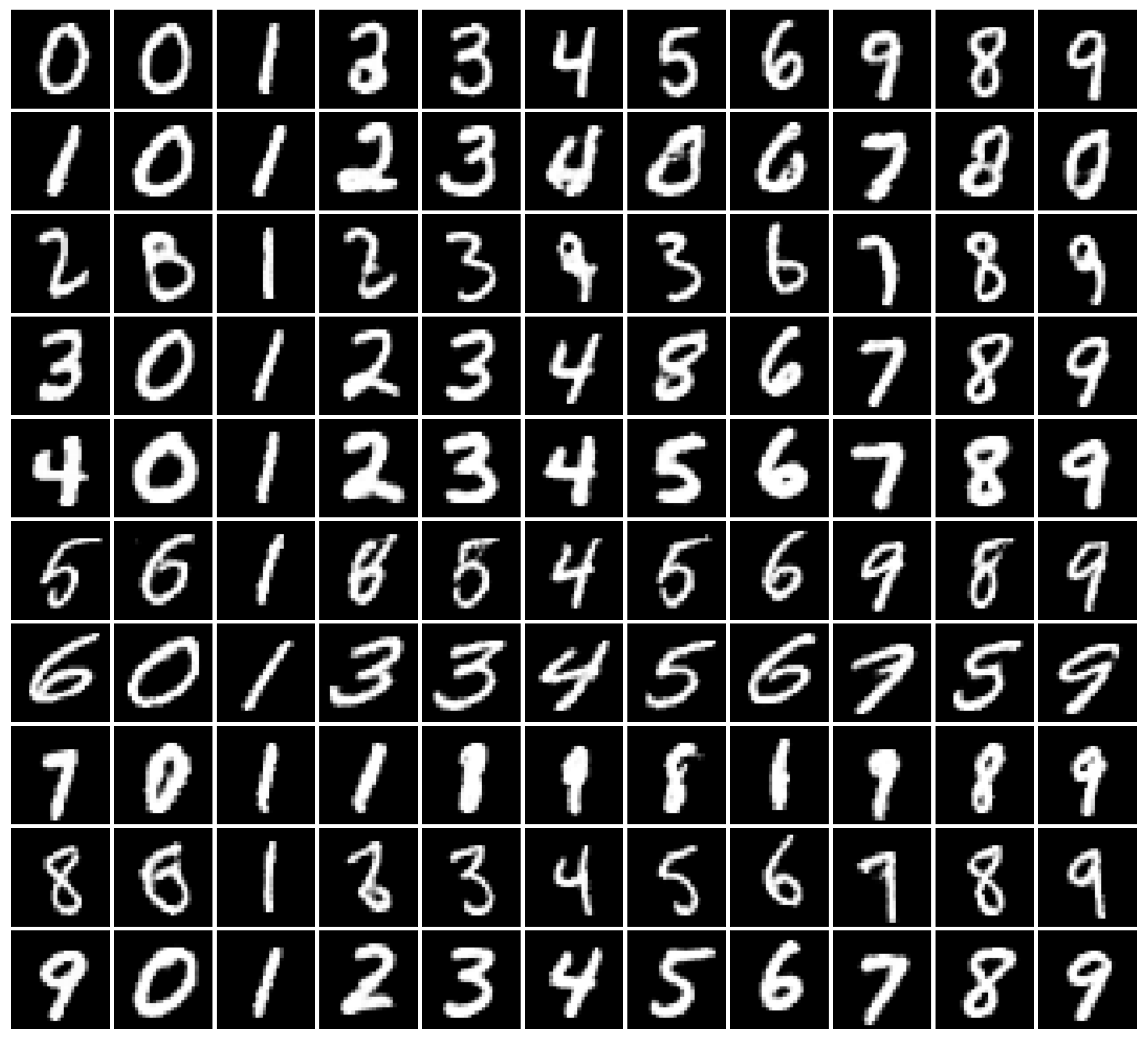}}\hspace{0.1cm}
	\subfloat[]{\includegraphics[width=0.19\textwidth]{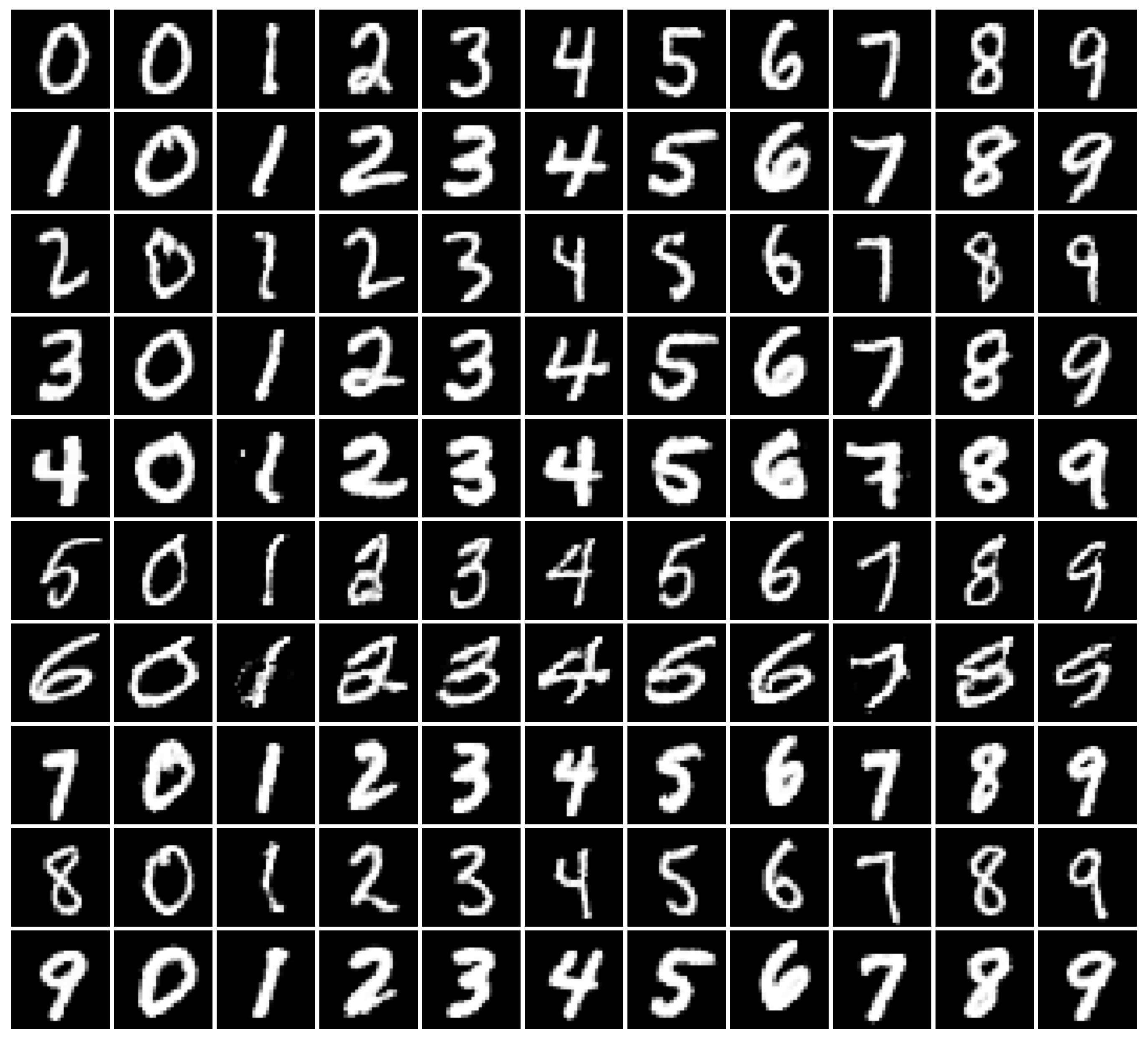}}\hspace{0.1cm}
	\subfloat[]{\includegraphics[width=0.19\textwidth]{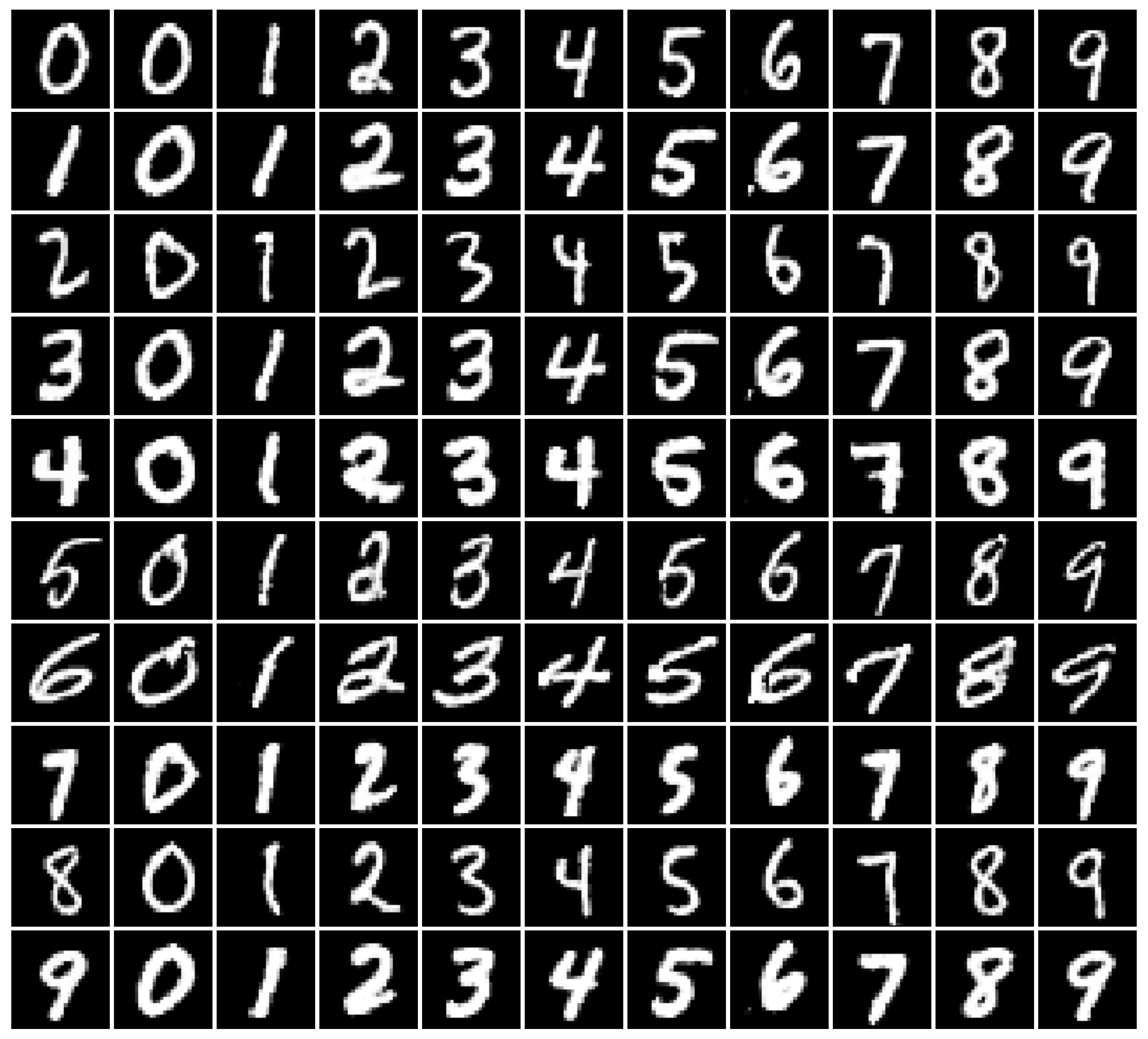}}%
	\caption{Generated digits with different handwriting styles. (a) AE-XCov. (b) IcGAN. (c) VAE/GAN. (d) CDNet-XCov. (e) CDNet-dCov. In each panel, the first column displays MNIST test images, and the other columns show analogical fantasies of test images.}\label{fig:generate_digits}
\end{figure*}

\begin{figure*}[!t]
	\centering
	\includegraphics[width=0.99\textwidth]{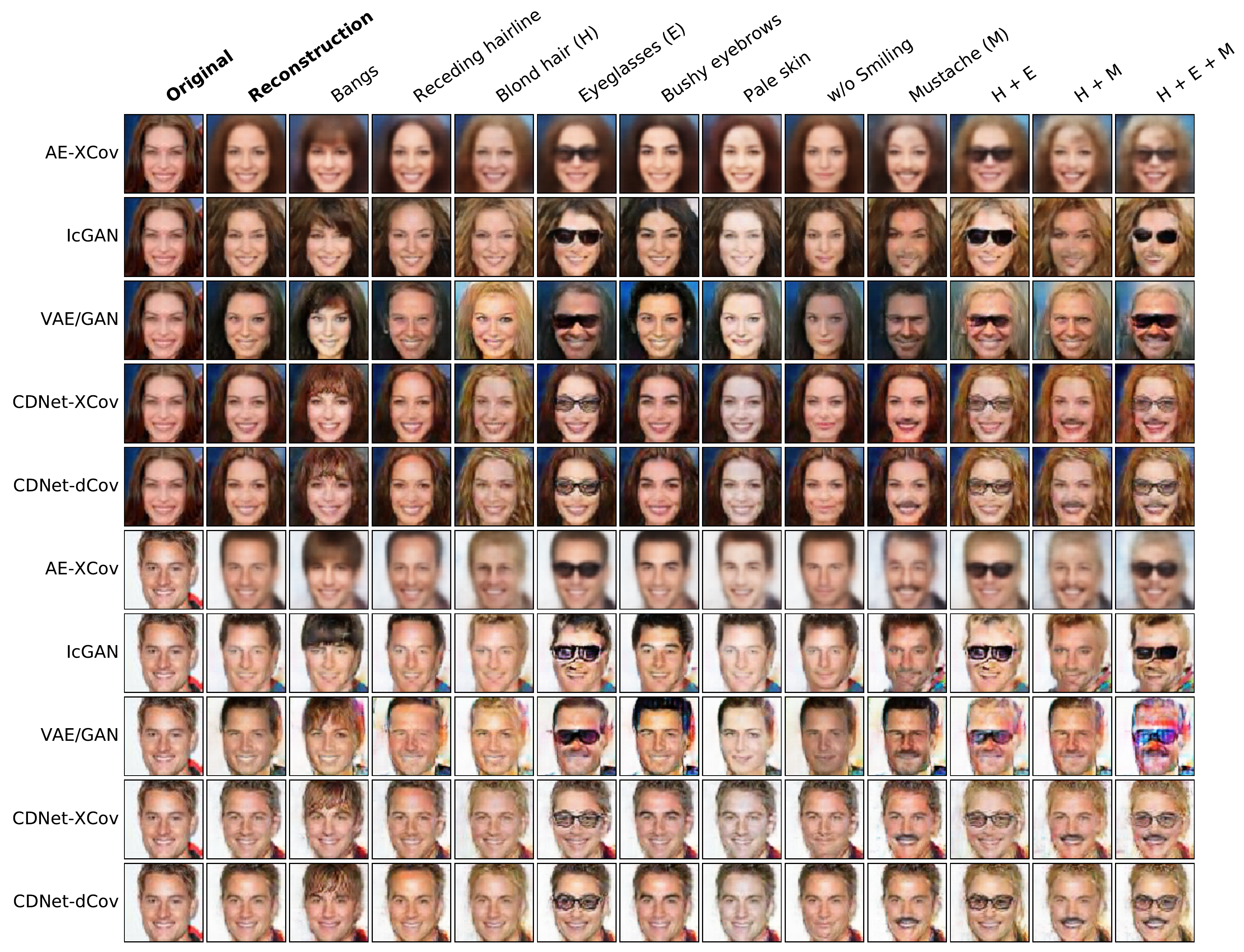}%
	\caption{Synthesized face images with the designated attributes. The first column shows original test images, the second column displays reconstructed counterparts, the next eight columns illustrate synthesized faces with the single target attribute, and the rightmost three columns correspond to examples of multi-attribute manipulations (H: Blond hair, E: Eyeglasses, M: Mustache). Best viewed in color.}\label{fig:generate_face_images}
\end{figure*}

In this group of experiments, we first use all models to generate new handwritten digits with the designated handwriting styles, such as boldface, italic, and broad shape. The image manipulation method is described as the \textit{Case 1} in Section \ref{sec:method_manipulate}. As shown in Fig. \ref{fig:generate_digits} (a), the AE-XCov presents a limited disentanglement ability, such as synthesizing new digits slightly leaning to the left (corresponding to the test digit ``2''). The disentanglement performance of the VAE/GAN is not stable, as many synthesized digits cannot reflect the category nature (see Fig. \ref{fig:generate_digits} (c)). We attribute this drawback to that in VAE/GAN, there still exist strong correlations between the learned representations of different digit classes. Both of the two CDNets, as well as the IcGAN, are able to generate new digits with the same style as originals, demonstrating the disentanglement of style from class.

\begin{figure*}[!t]
	\centering
	\subfloat[]{\includegraphics[width=0.33\textwidth]{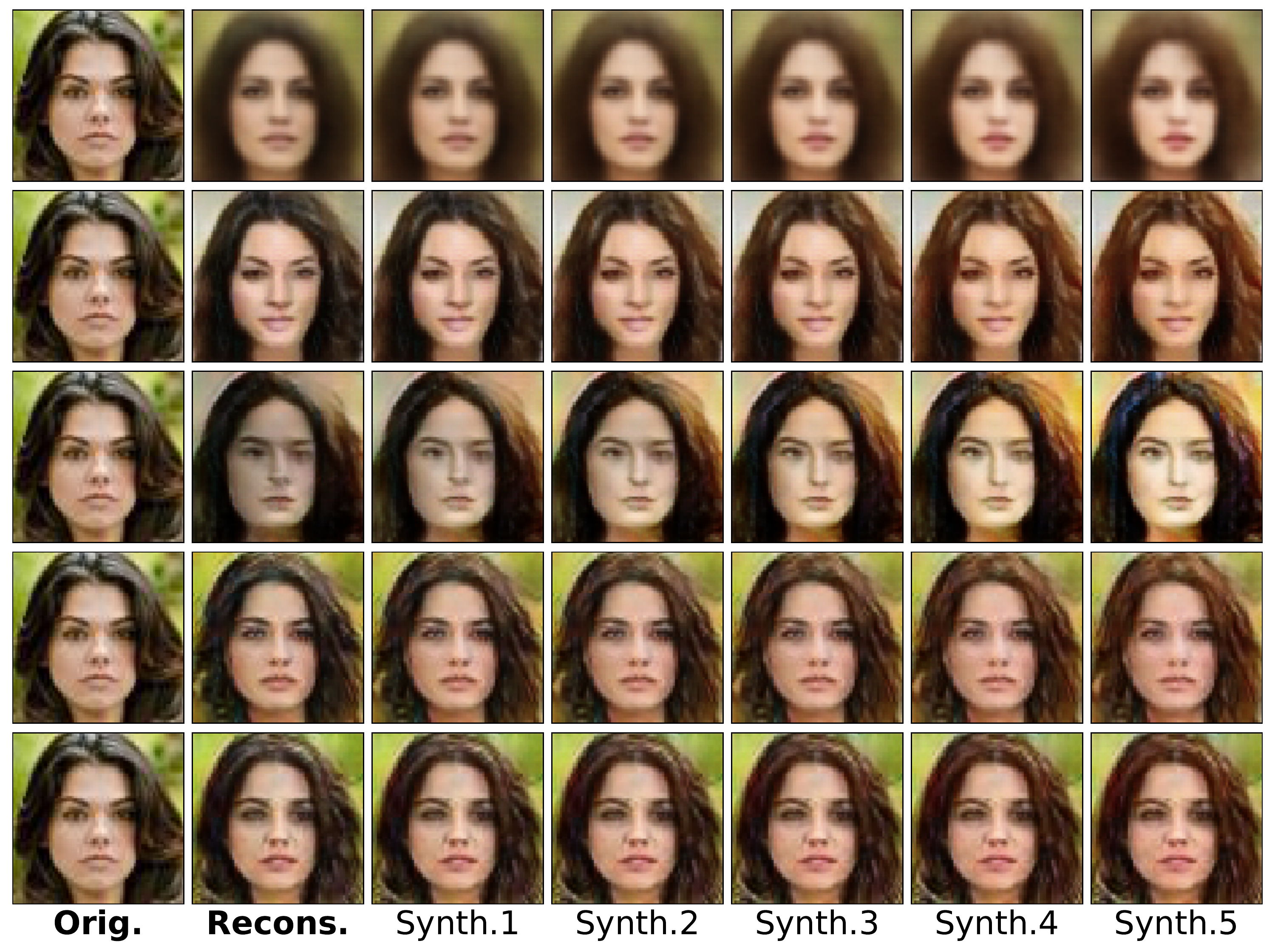}}
	\subfloat[]{\includegraphics[width=0.33\textwidth]{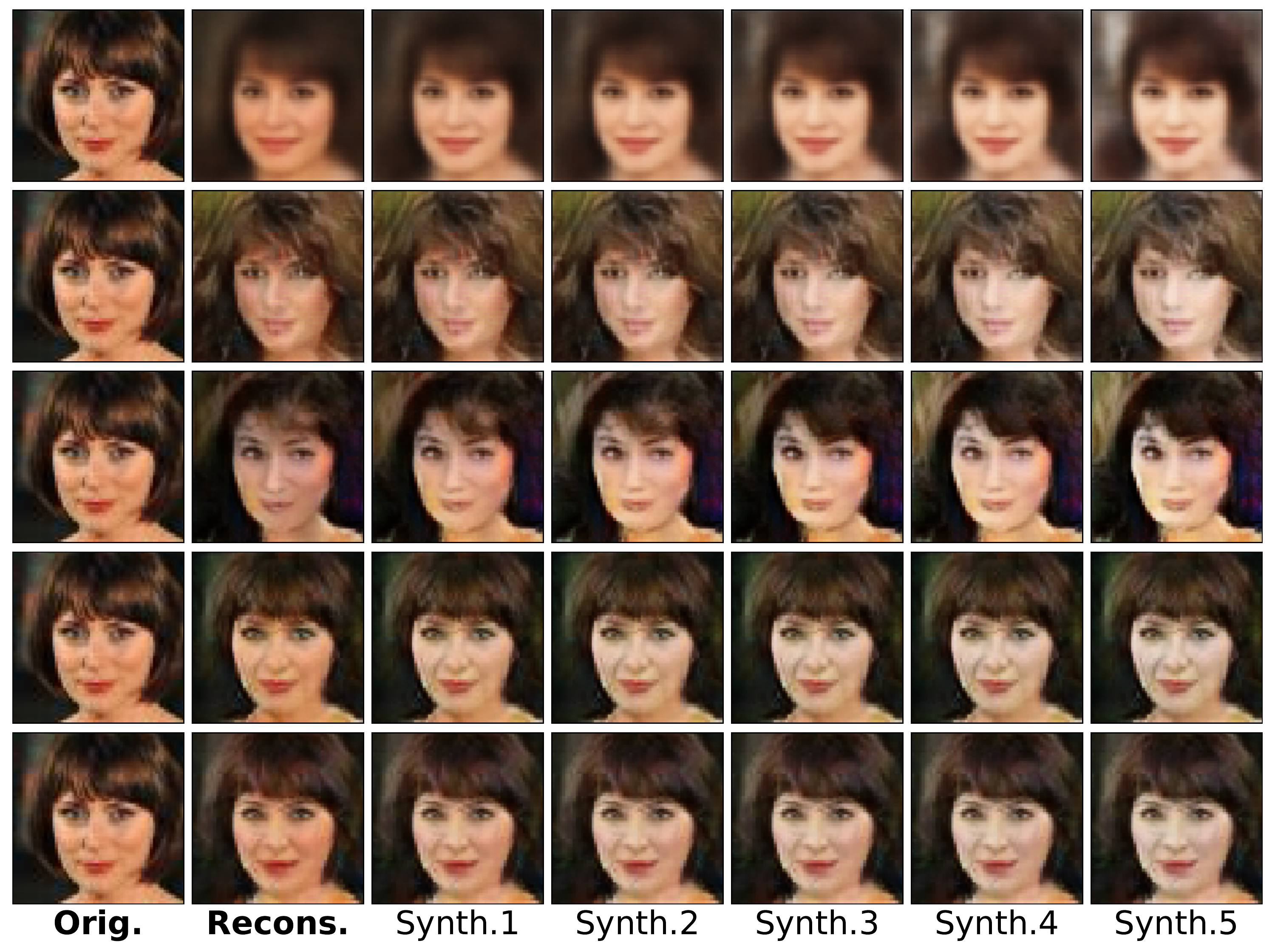}}
	\subfloat[]{\includegraphics[width=0.33\textwidth]{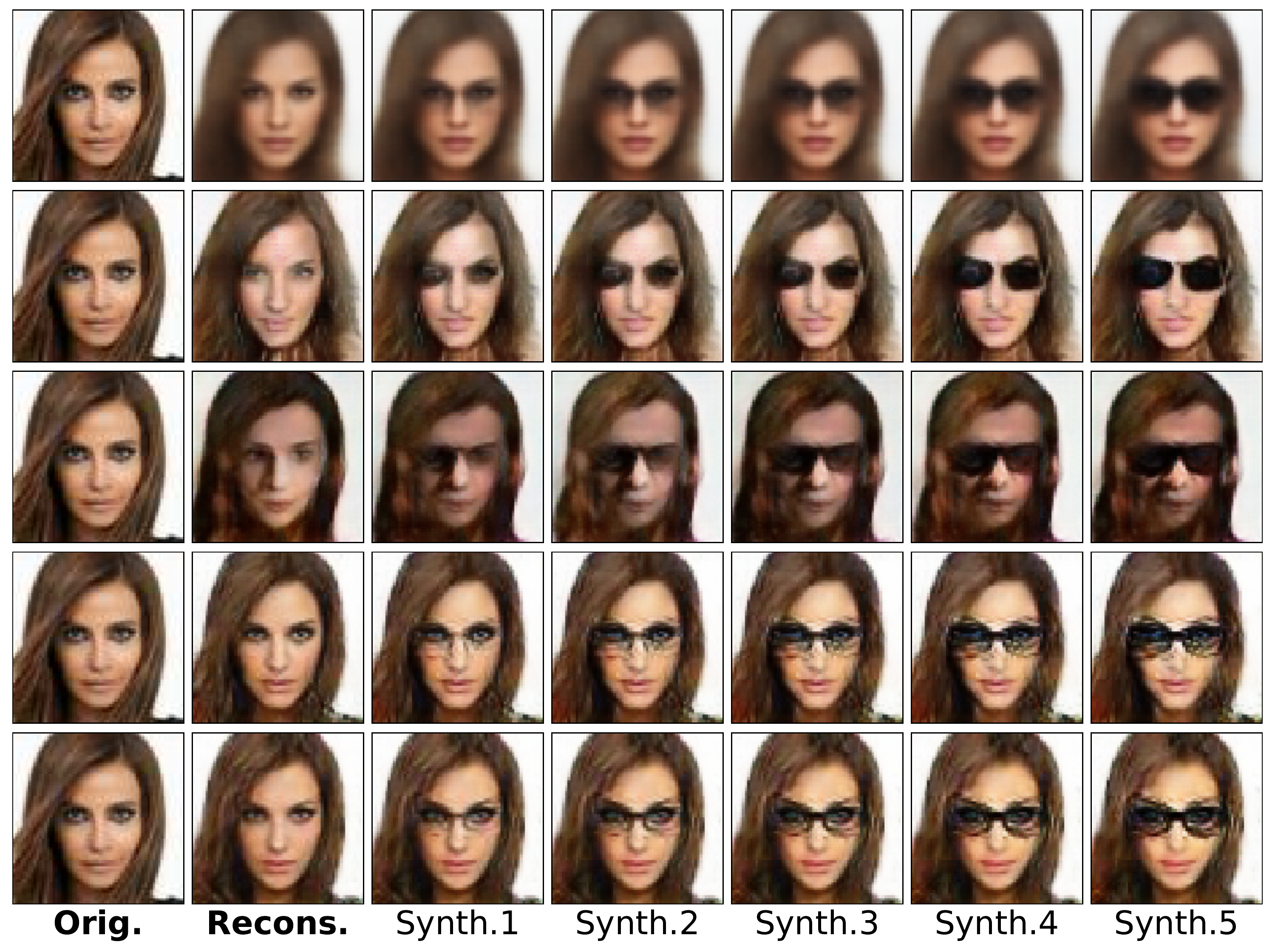}}\\ \vspace{-2mm}
	\subfloat[]{\includegraphics[width=0.33\textwidth]{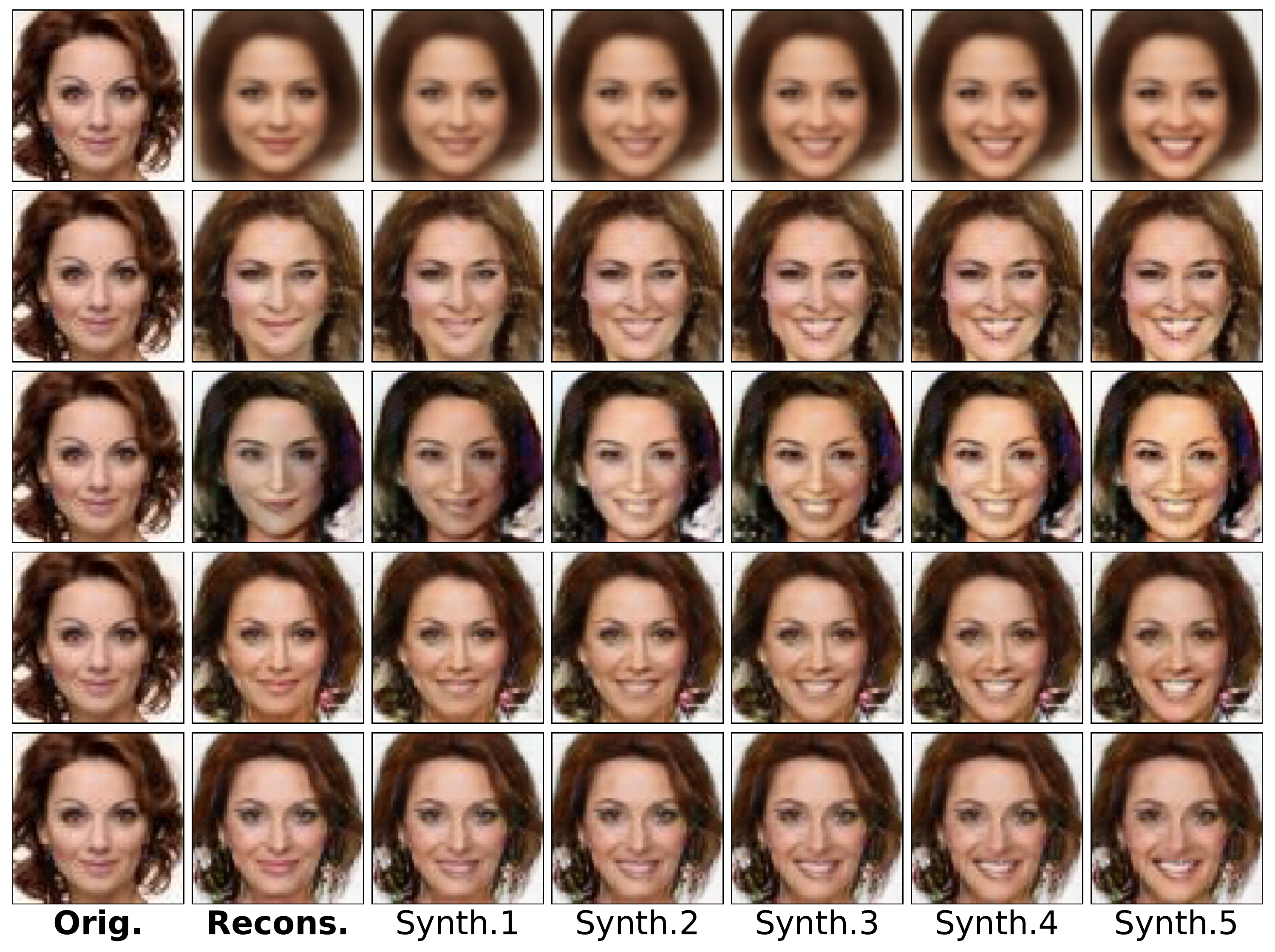}}
	\subfloat[]{\includegraphics[width=0.33\textwidth]{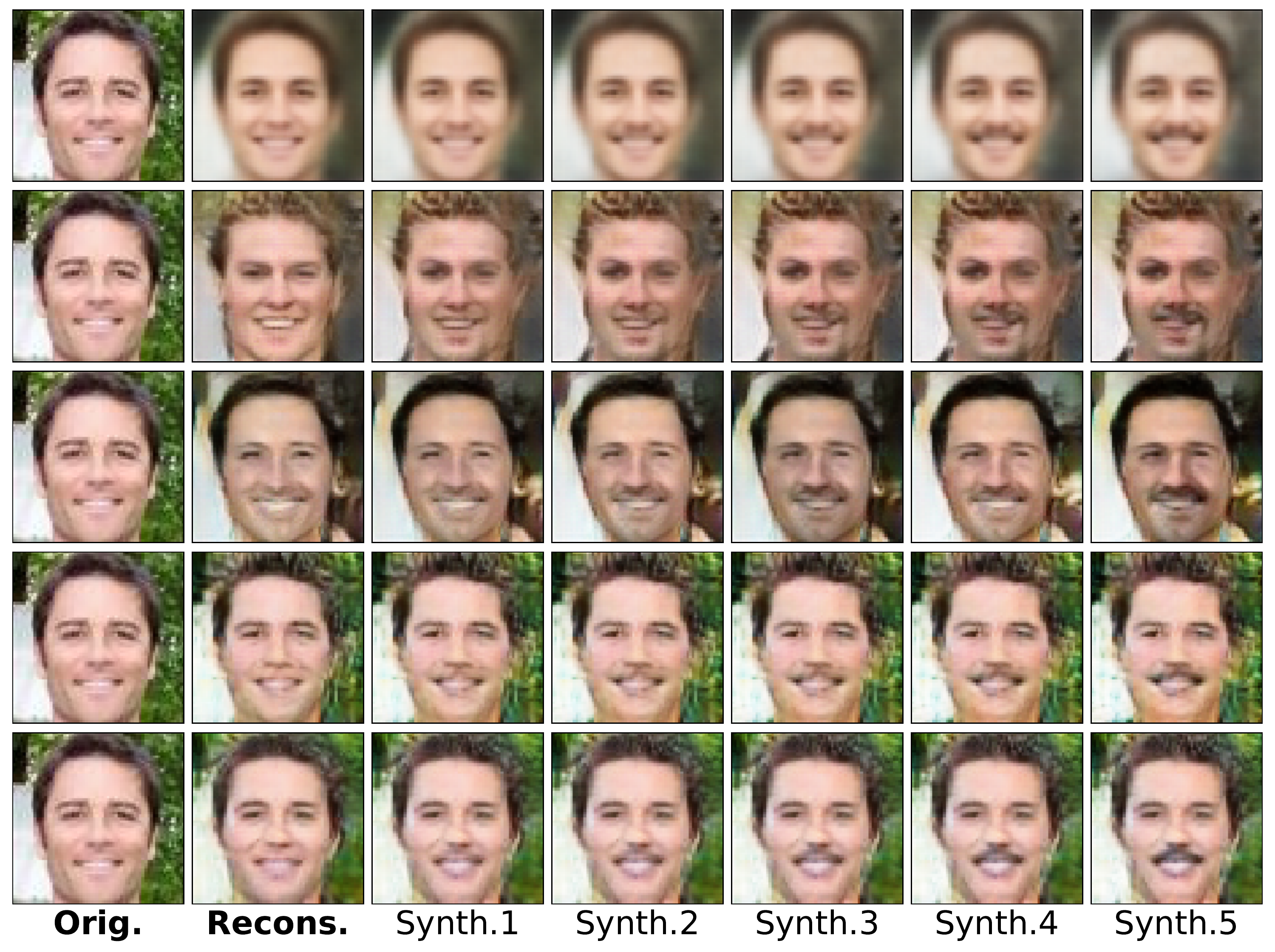}}
	\subfloat[]{\includegraphics[width=0.33\textwidth]{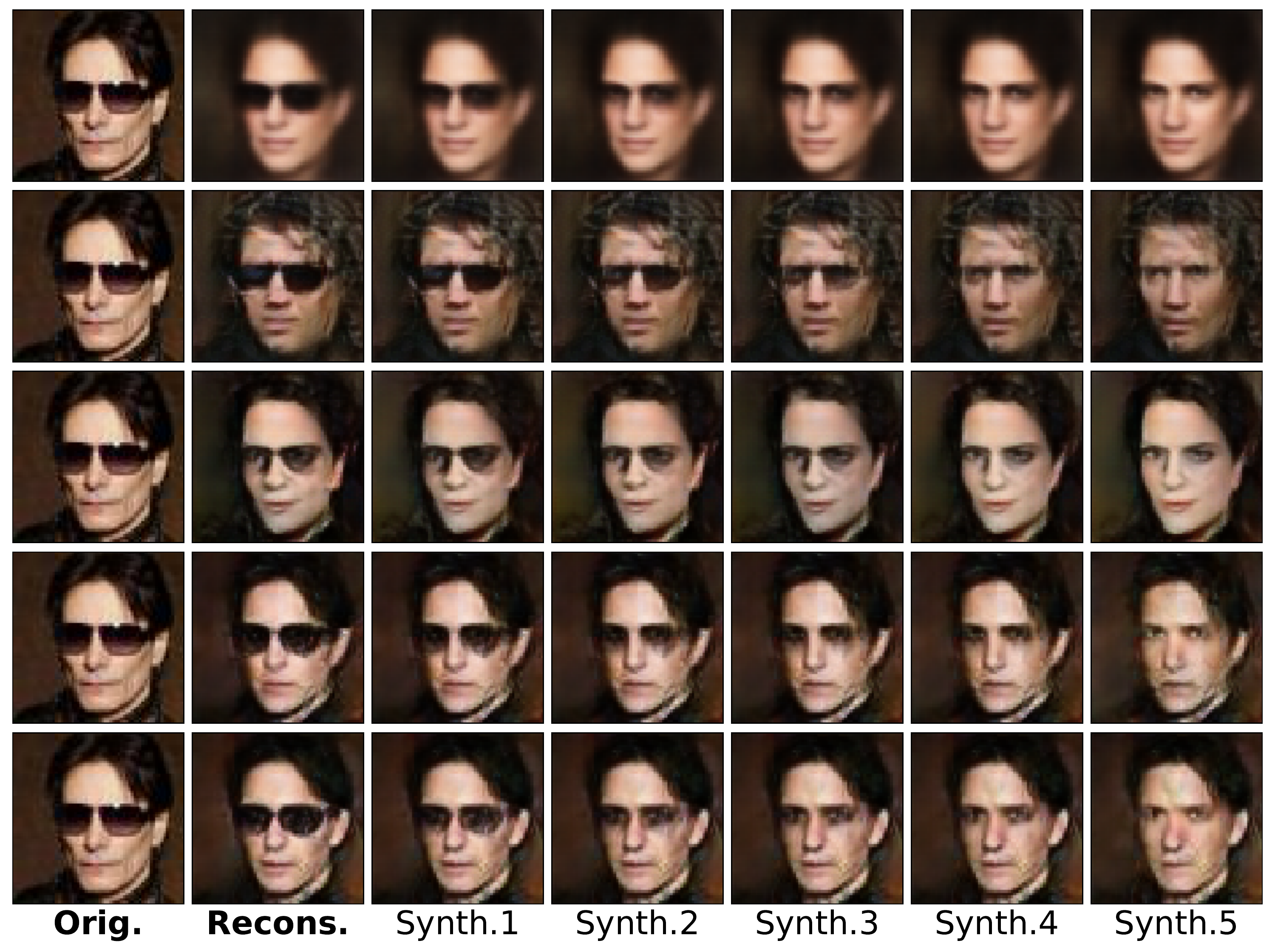}}%
	\caption{Synthesized face images with various attribute intensities. (a) Brown hair. (b) Pale skin. (c) Eyeglasses. (d) Smiling. (e) Mustache. (f) w/o Eyeglasses. The results in each panel, from the first row to the last row, are obtained by AE-XCov, IcGAN, VAE/GAN, CDNet-XCov, and CDNet-dCov, respectively. In each panel, the first column shows the original test image, the second column for reconstructions, and the remaining five columns for synthesized images with different attribute intensities, from weaker levels to stronger ones. Best viewed in color.}\label{fig:generate_face_images_diff_degrees}
\end{figure*}

Second, we aim to synthesize new faces with the specific facial attributes while preserving the core identity. The manipulation method is described as the \textit{Case 2} in Section \ref{sec:method_manipulate}. And the disentanglement performance of different models on CelebA face images is illustrated in Fig. \ref{fig:generate_face_images}. We can observe that the AE-XCov can produce new faces with the desired attributes, but the blurring effects obviously degrade the output images' visual quality. For the IcGAN model, although target attributes are clearly involved in the resulting images, the core object identities have been changed as occurred in the reconstruction task. The synthetic results of VAE/GAN turn out to be really sensitive to the attribute modifications. In particular, when leveraging the VAE/GAN to add the ``Eyeglasses'' or ``Mustache'' attribute into the original female face, the hairstyle and even the gender of the subject are apparently altered. When it comes to the multi-attribute manipulations, all three baseline models fail to mix different attributes at once, since the multi-attribute changes induce the problem of visually-perceptible image distortion. By contrast, the two CDNets exhibit a remarkable ability to disentangle facial attributes from identity, that is, all designated attributes are incorporated into the final images in a more natural manner.

It's worth noting that on the multi-attribute manipulation task, especially for the male face image editing, the CDNet-dCov visually outperforms the CDNet-XCov in terms of preserving the core identity information (e.g., see deformations of the mouth area shown in the last three columns of Fig. \ref{fig:generate_face_images}). We will give a quantitative comparison between them in Section \ref{sec:disentangle_class}.

\subsection{Controllable Disentanglement}\label{sec:disentangle_degree}
In this experiment, we \textit{qualitatively} compare the disentanglement strength of all baselines and the CDNet by synthesizing face images with various attribute intensities. The image manipulation method is similar to the \textit{Case 2} in Section \ref{sec:method_manipulate}, and here we take a list of values to orderly modify the corresponding attribute representations. The attribute value range is set to $[-0.5, 6.0]$ for AE-XCov, $[0.0, 6.0]$ for IcGAN, $[-1.0, 3.0]$ for VAE/GAN, and $[-1.5, 3.0]$ for CDNet. We conduct this group of experiments on six representative facial attributes, i.e., ``Brown hair'', ``Pale skin'', ``Eyeglasses'', ``Smiling'', ``Mustache'', and ``w/o Eyeglasses'', respectively. From Fig. \ref{fig:generate_face_images_diff_degrees}, it is observed that the blurring effect still exists across all synthetic images generated by AE-XCov. The IcGAN performs well to synthesize a set of new faces with target attributes and attribute intensities; however, none of the resulting images can preserve the core object identities very well. For the VAE/GAN, changing one attribute usually causes the deformation of other attributes. One of such examples is that the modifications on ``Pale skin'' attribute also cause the deformation of hairstyle. Our two CDNet models, by contrast, are competent to generate distinguishable face fantasies across all compared attributes and variation degrees, and meanwhile preserving the core identity. These results illustrate that the learning strategy of representations, as well as the image manipulation method, enables CDNet to control the degree of disentanglement during image editing.

\subsection{Comparing Disentanglement by Classification}\label{sec:disentangle_class}
\begin{figure*}[!t]
	\centering
	\subfloat[]{\includegraphics[width=0.20\textwidth]{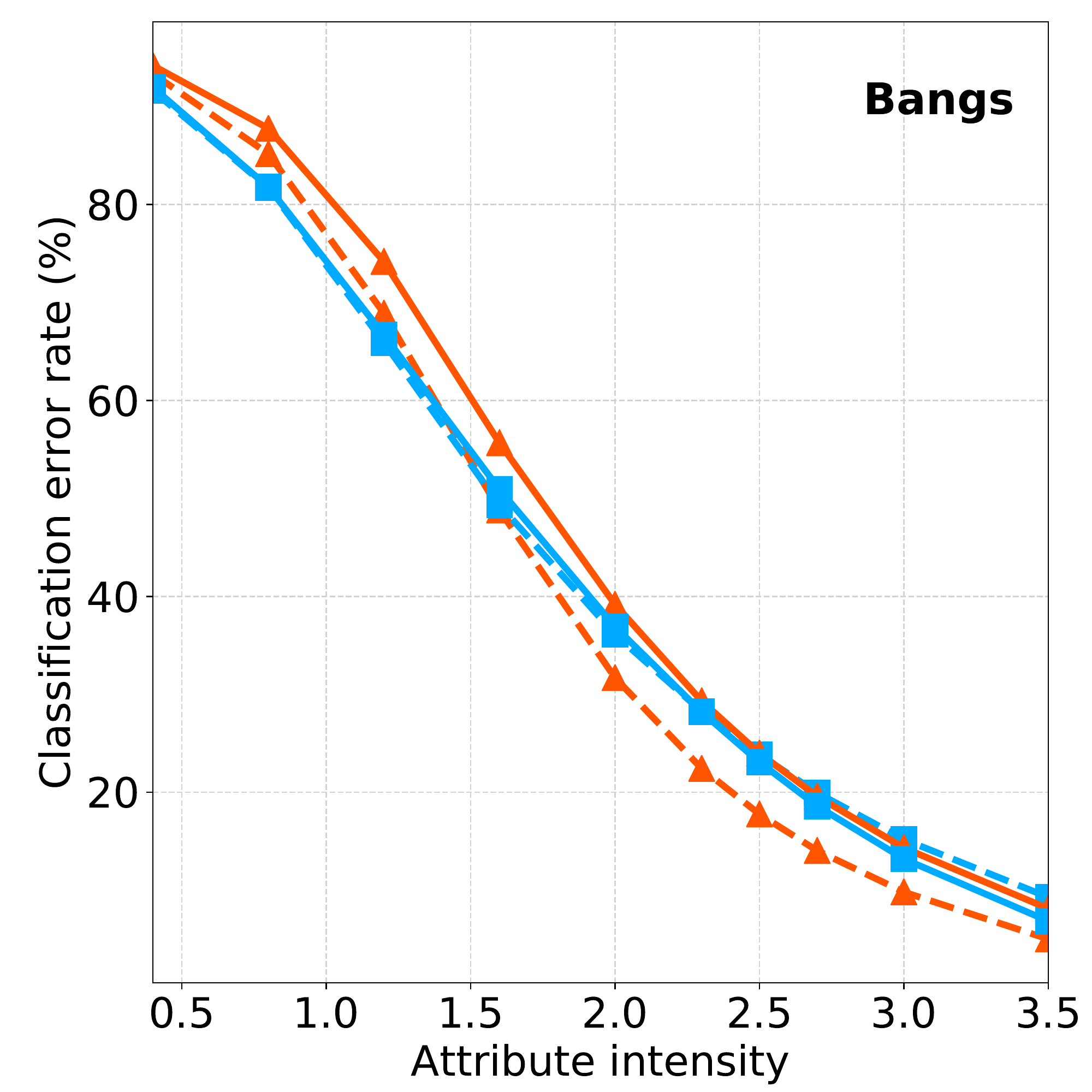}}
	\subfloat[]{\includegraphics[width=0.20\textwidth]{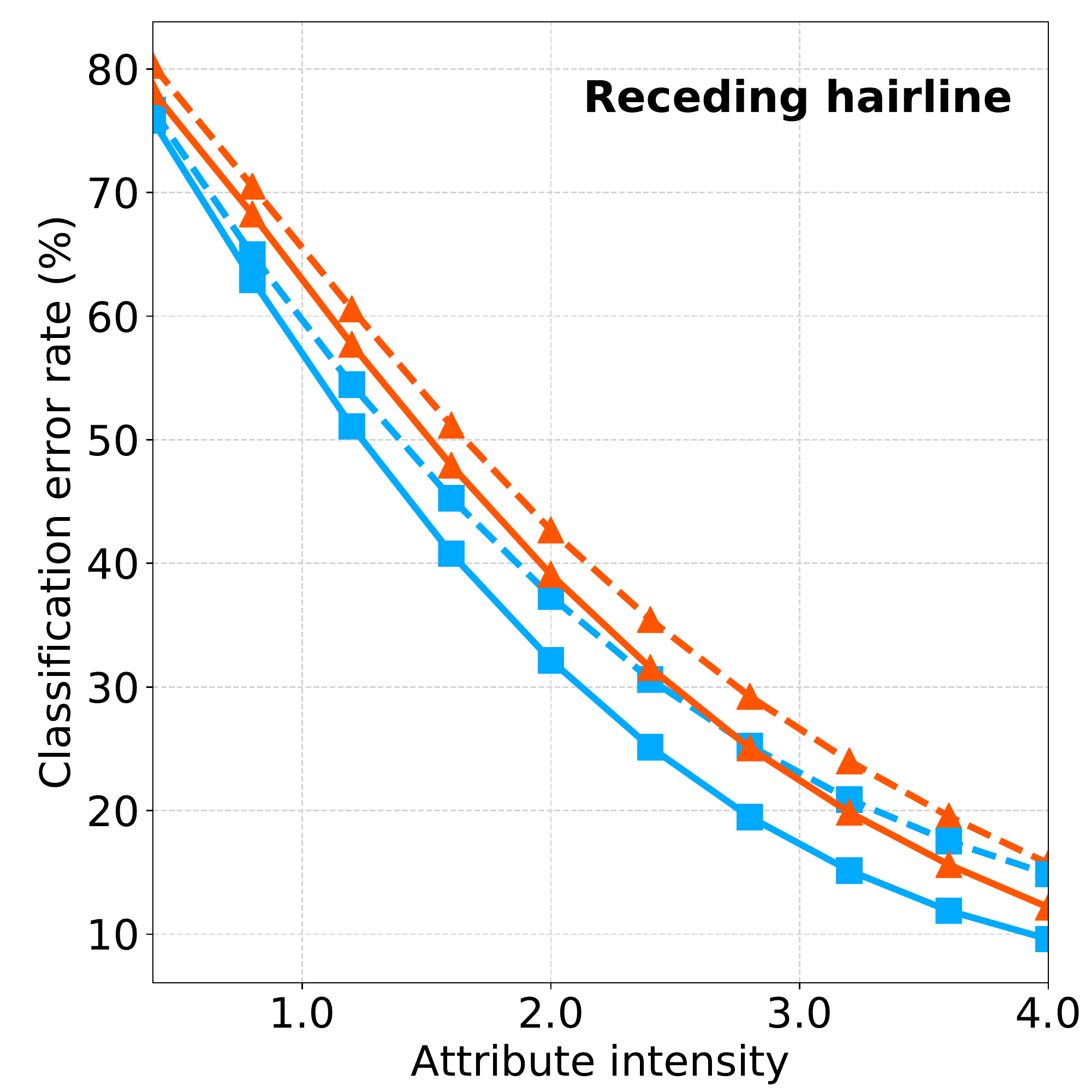}}
	\subfloat[]{\includegraphics[width=0.20\textwidth]{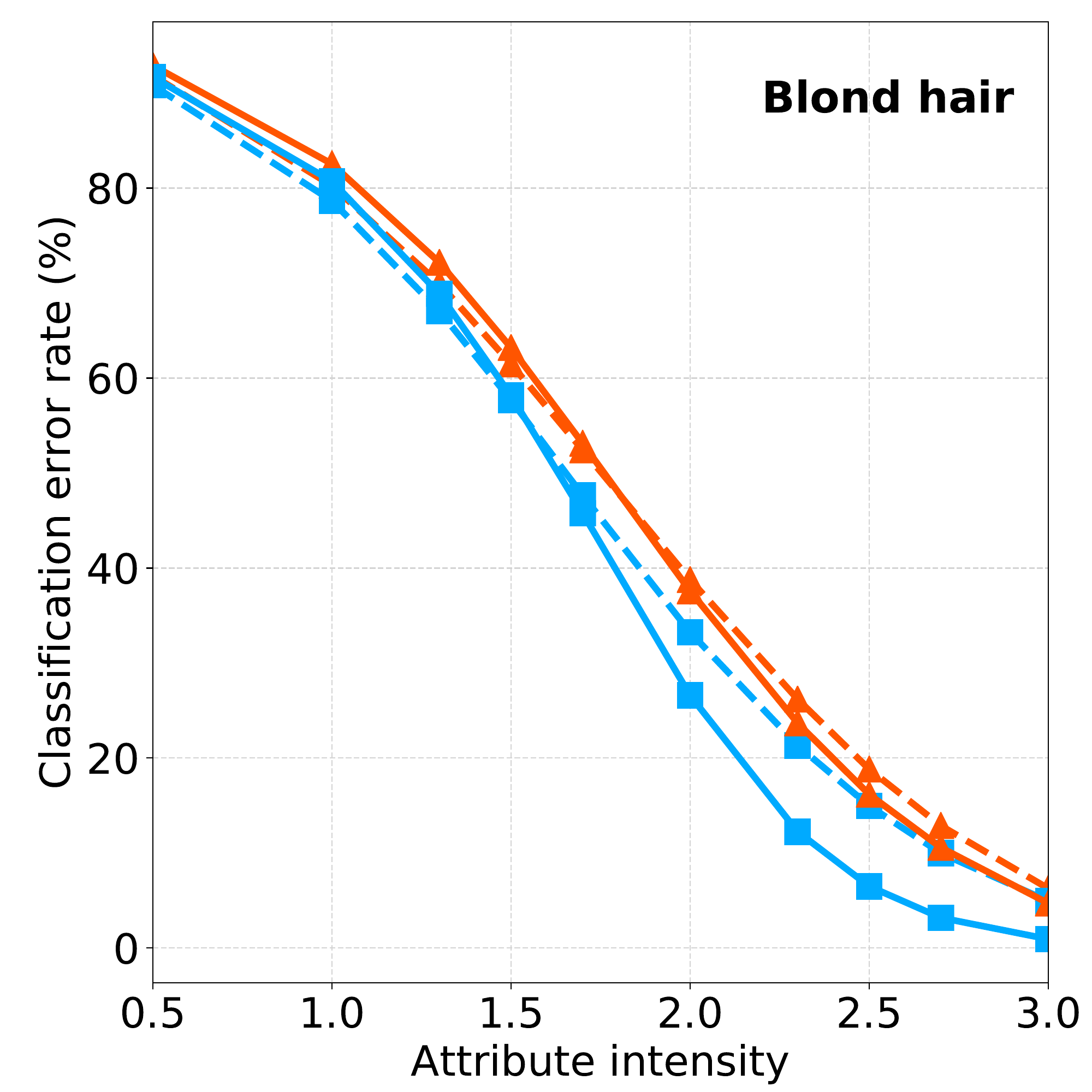}}
	\subfloat[]{\includegraphics[width=0.20\textwidth]{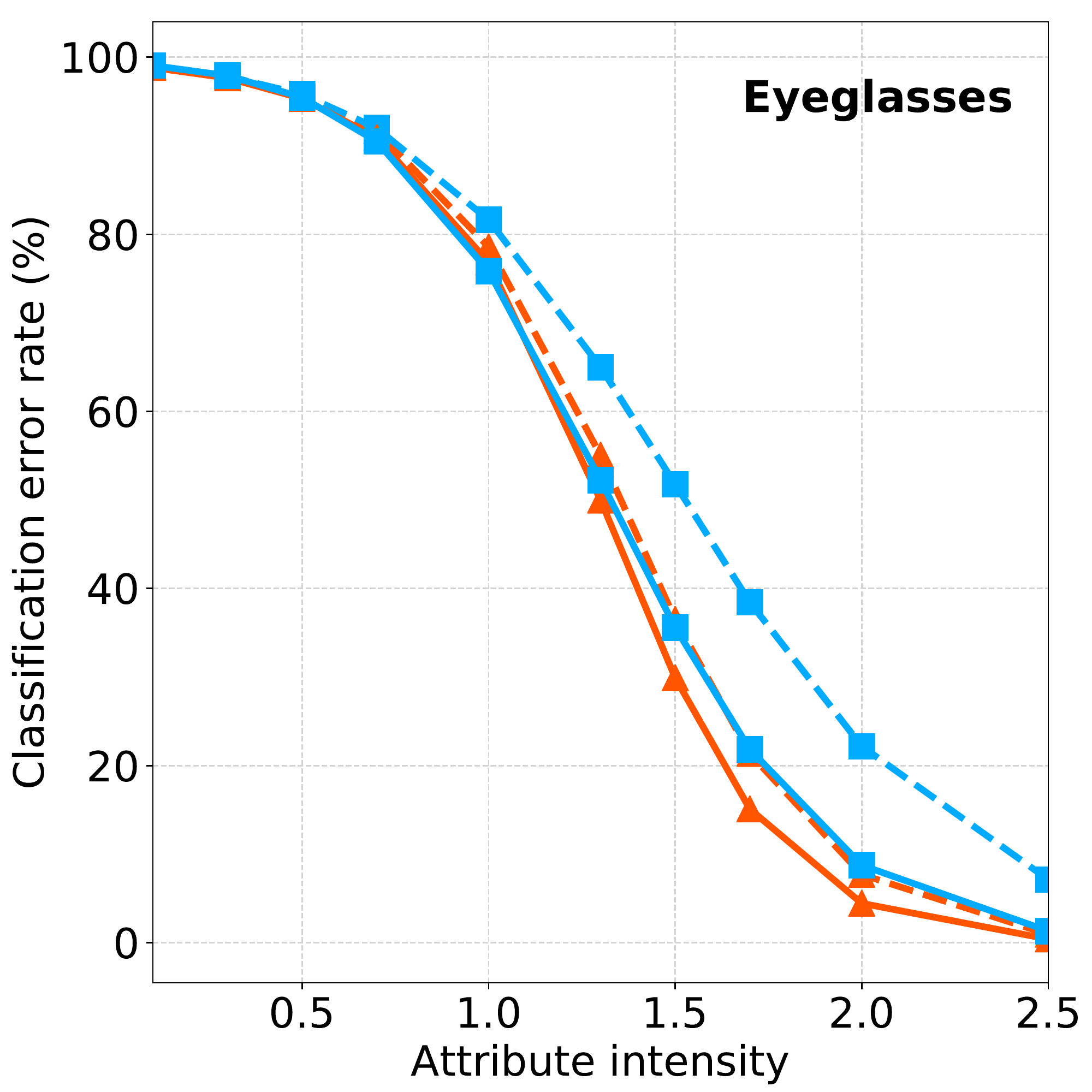}}
	\subfloat[]{\includegraphics[width=0.20\textwidth]{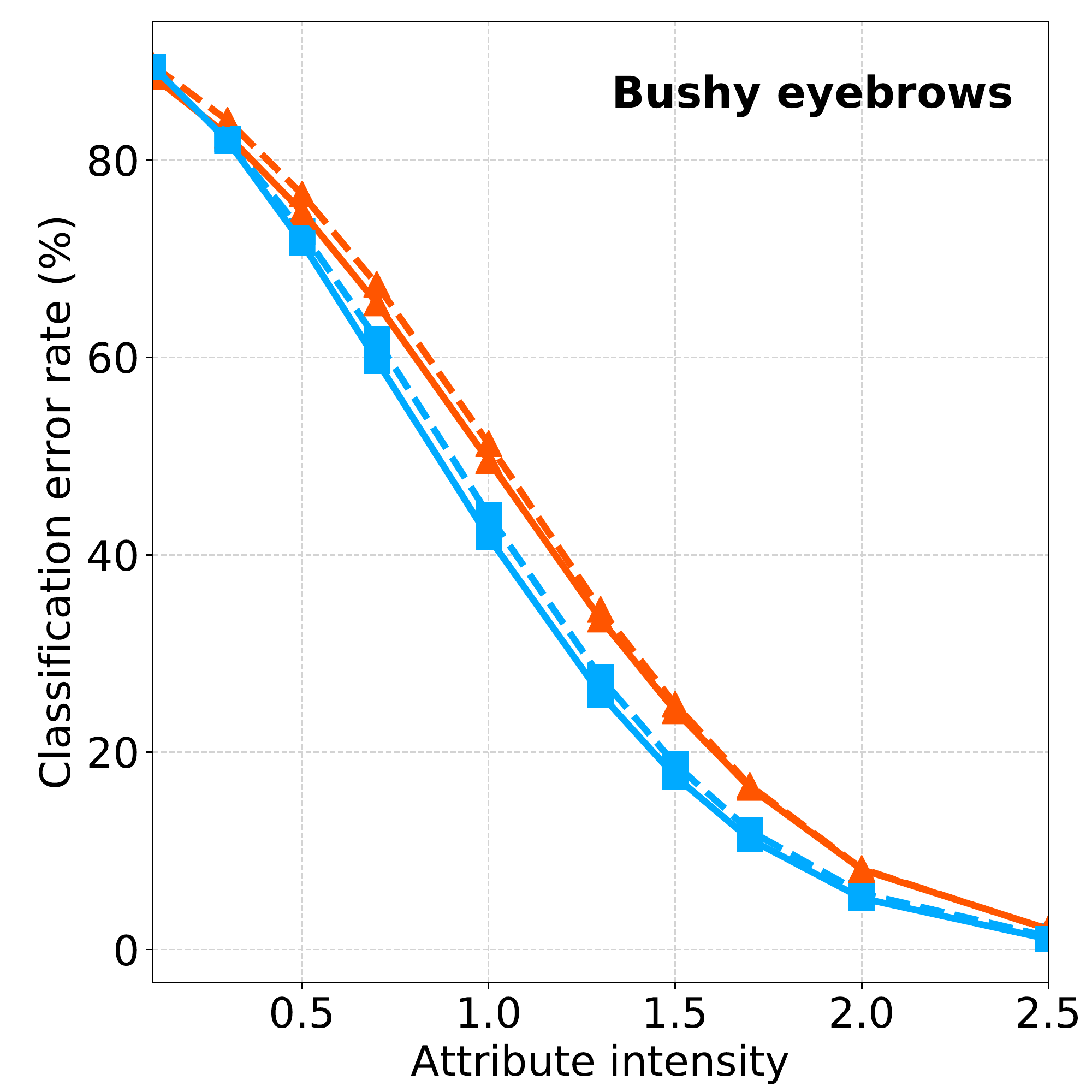}}\\ \vspace{-2mm}
	\subfloat[]{\includegraphics[width=0.20\textwidth]{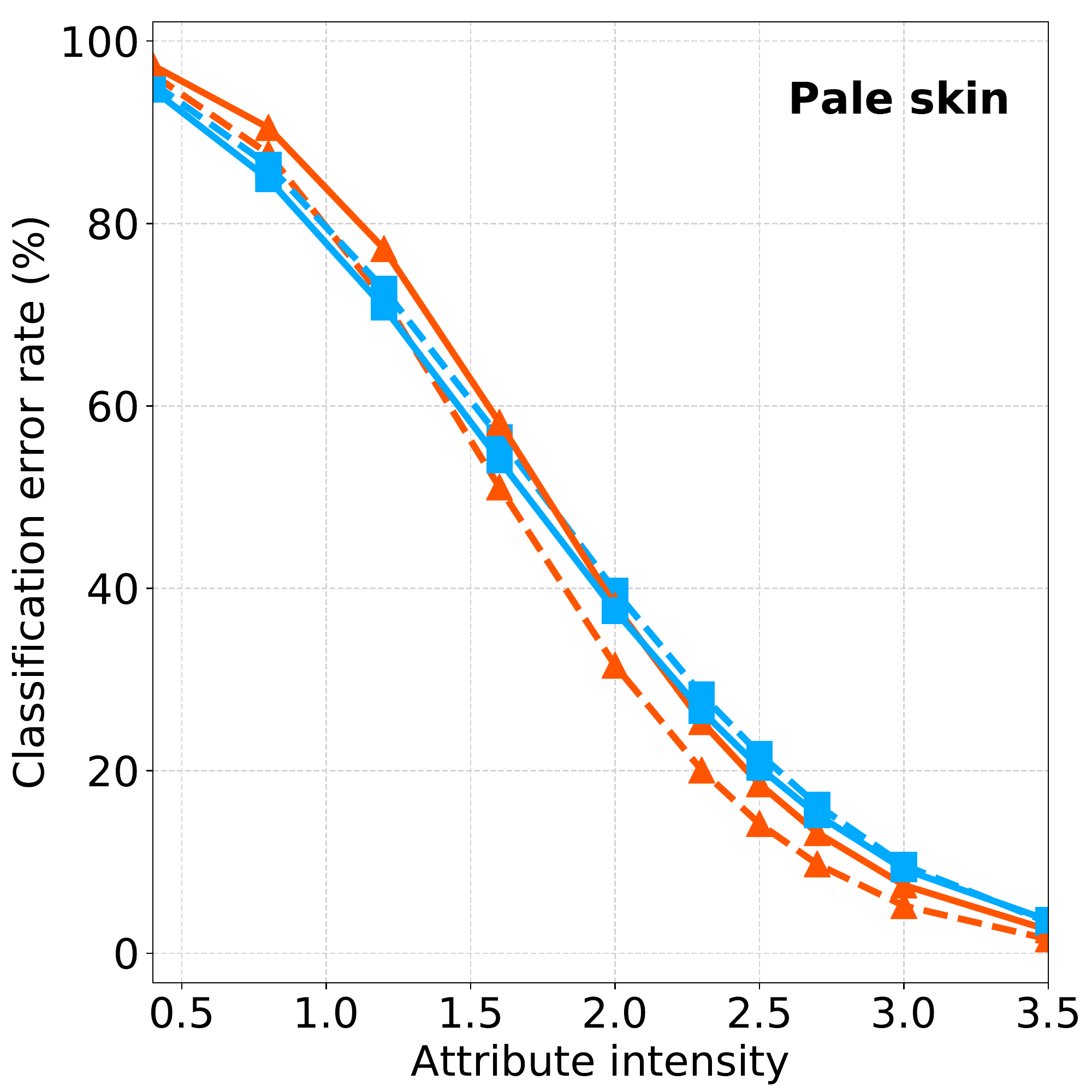}}
	\subfloat[]{\includegraphics[width=0.20\textwidth]{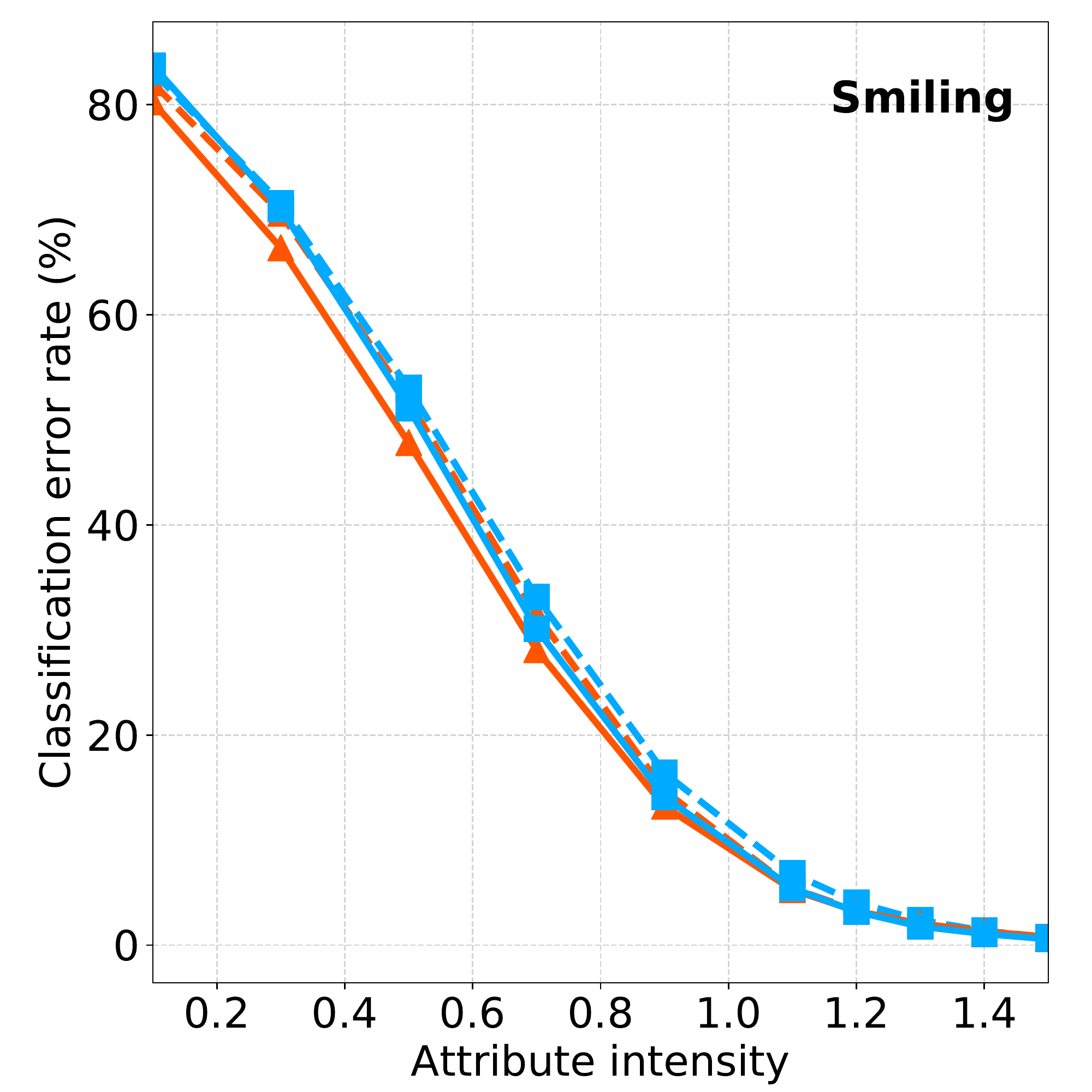}}
	\subfloat[]{\includegraphics[width=0.20\textwidth]{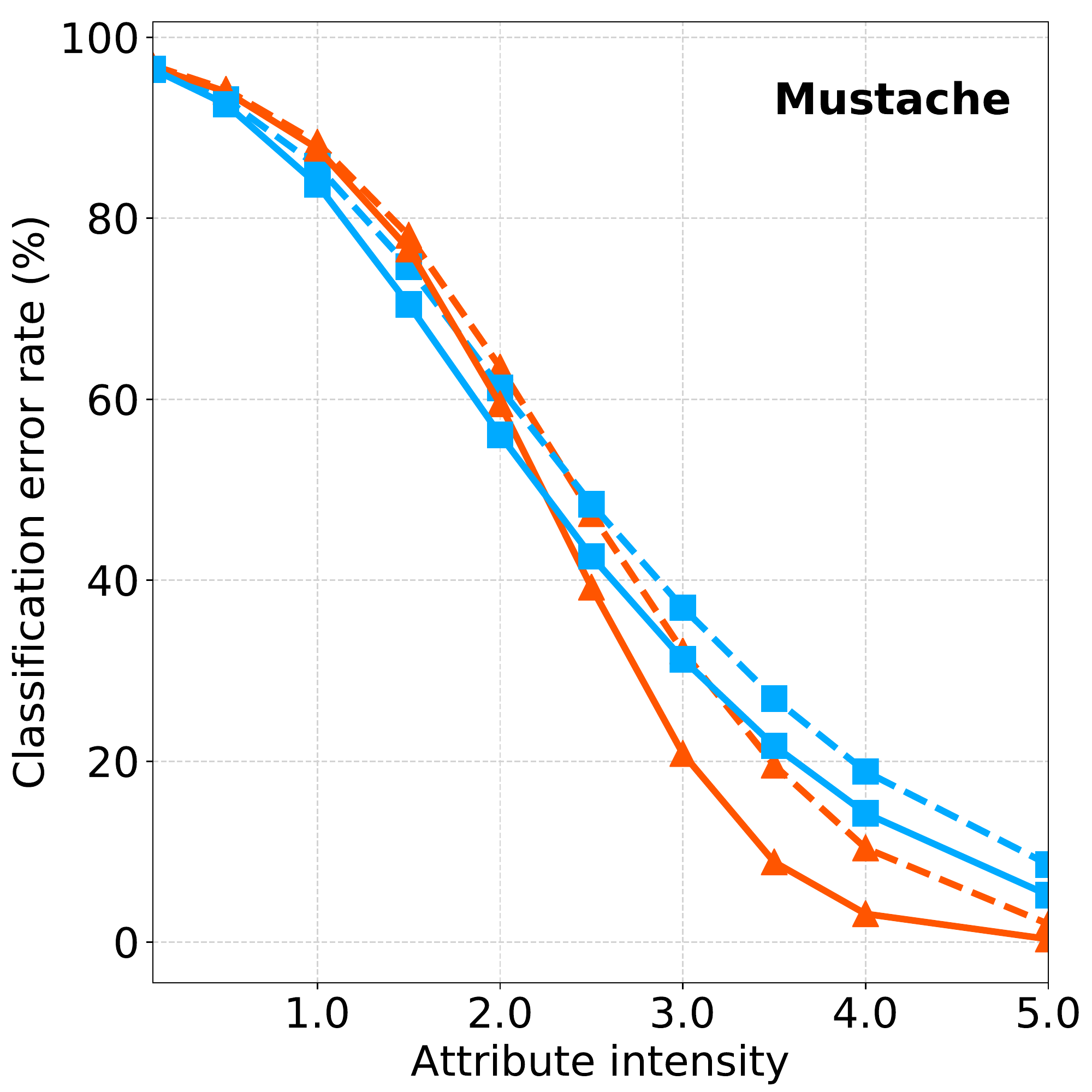}}
	\subfloat[]{\includegraphics[width=0.20\textwidth]{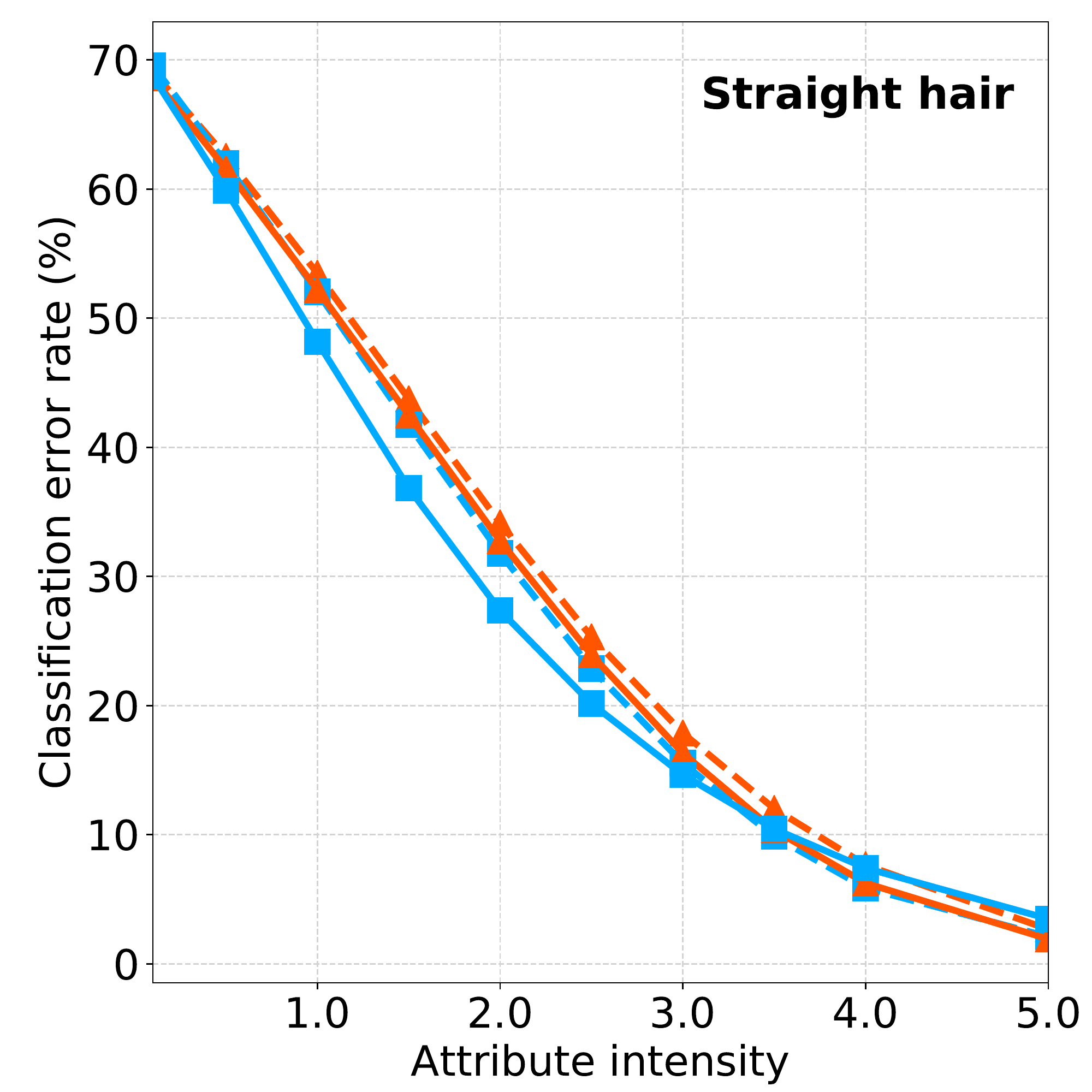}}
	\subfloat[]{\includegraphics[width=0.20\textwidth]{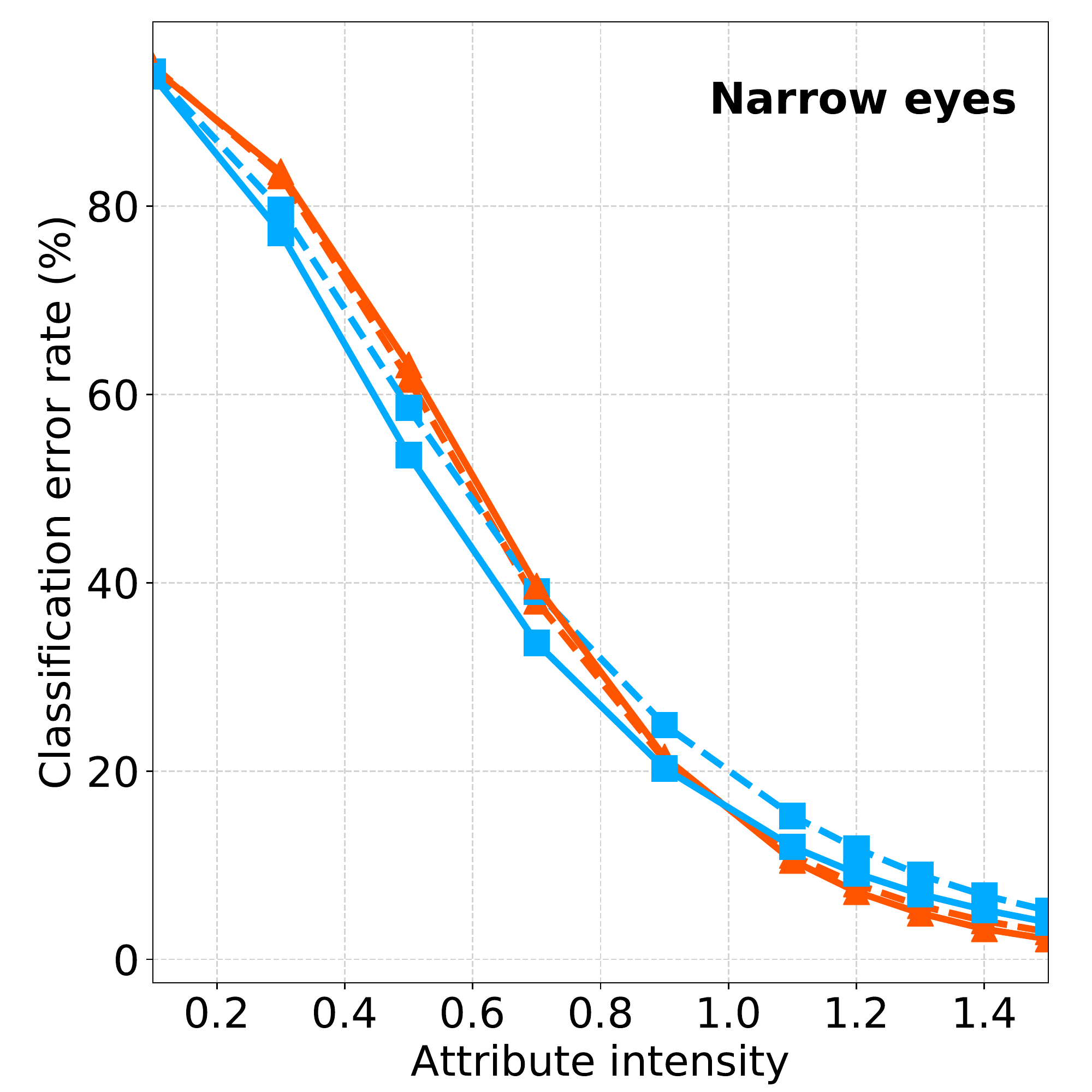}}\\ \vspace{-2mm}
	\subfloat[]{\includegraphics[width=0.20\textwidth]{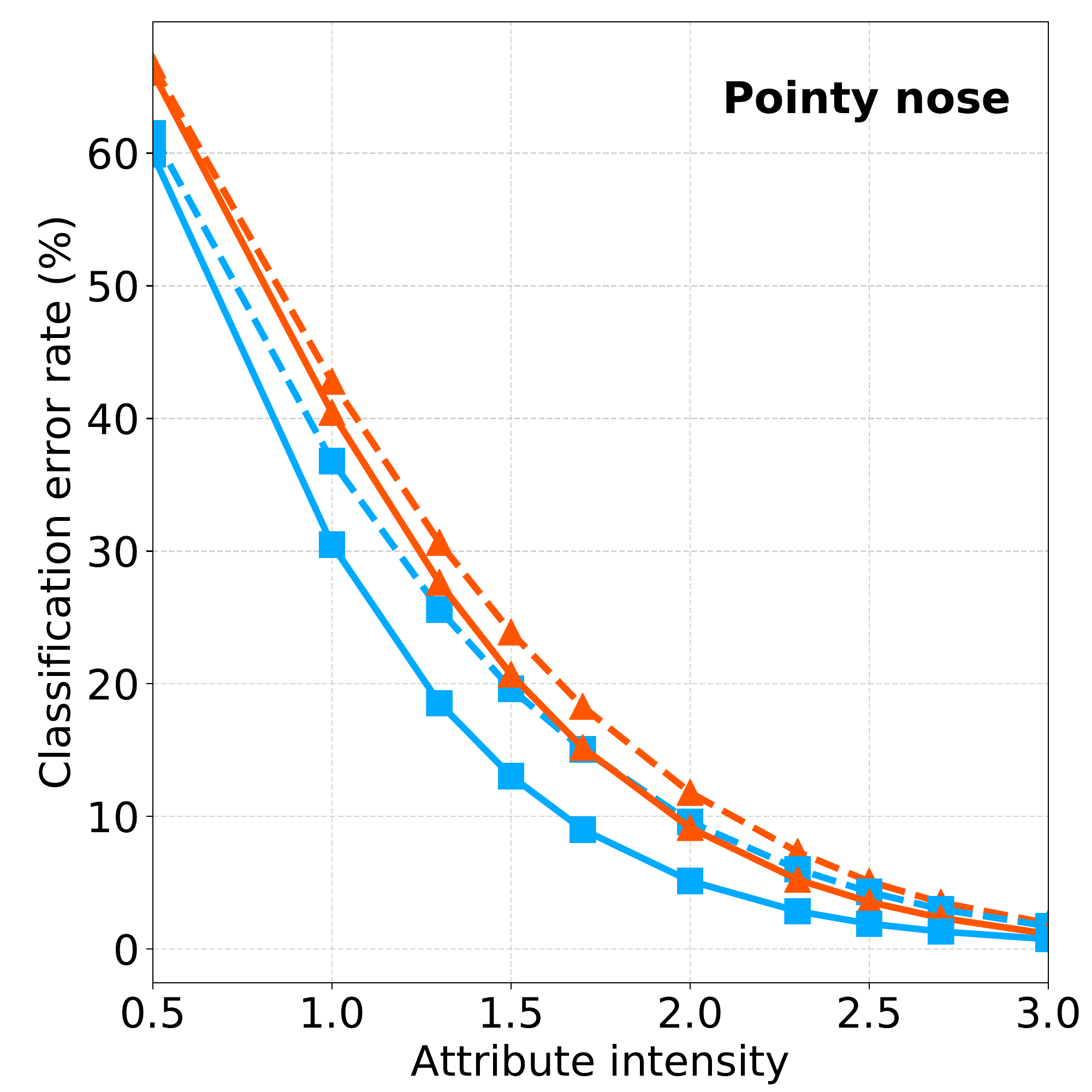}}
	\subfloat[]{\includegraphics[width=0.20\textwidth]{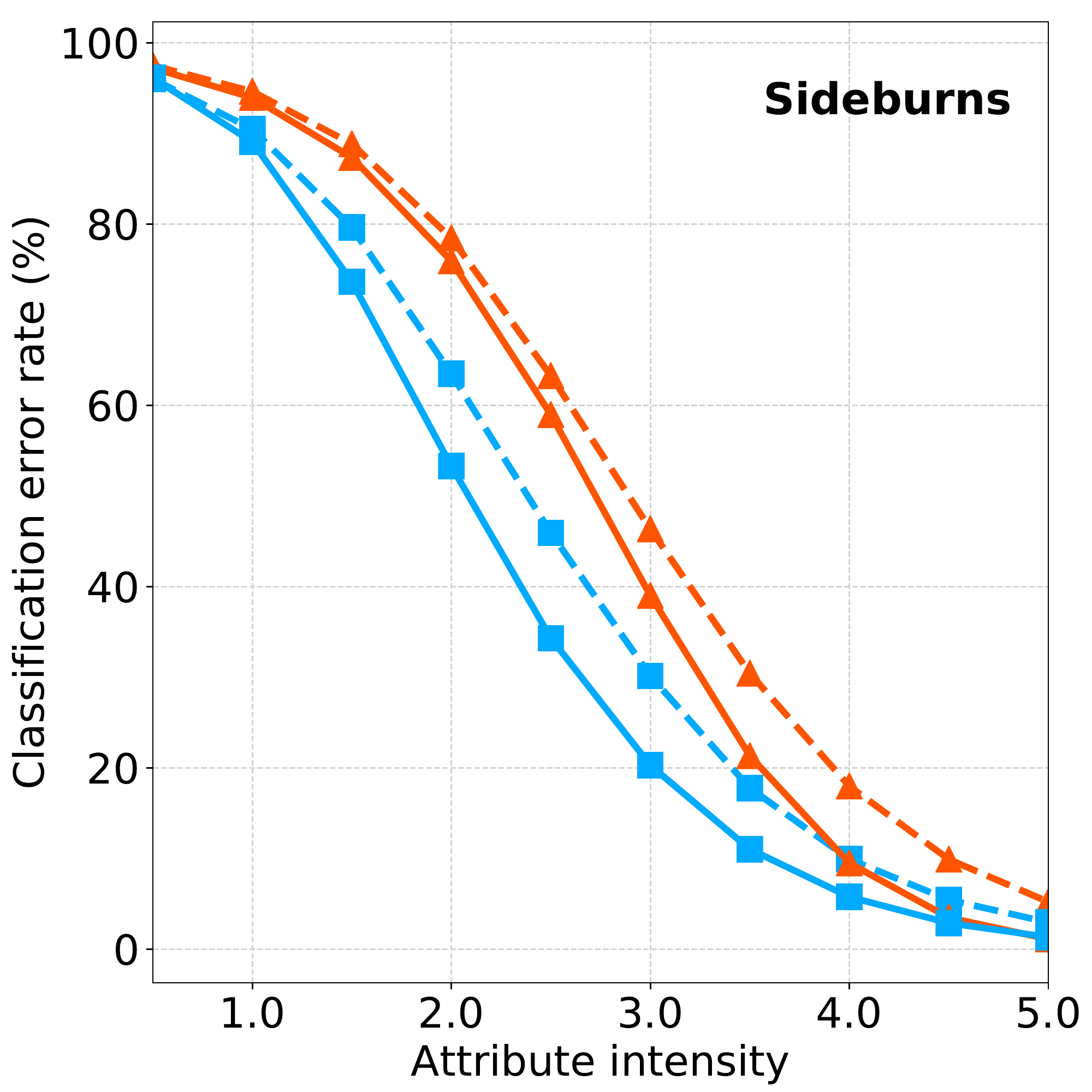}}
	\subfloat[]{\includegraphics[width=0.20\textwidth]{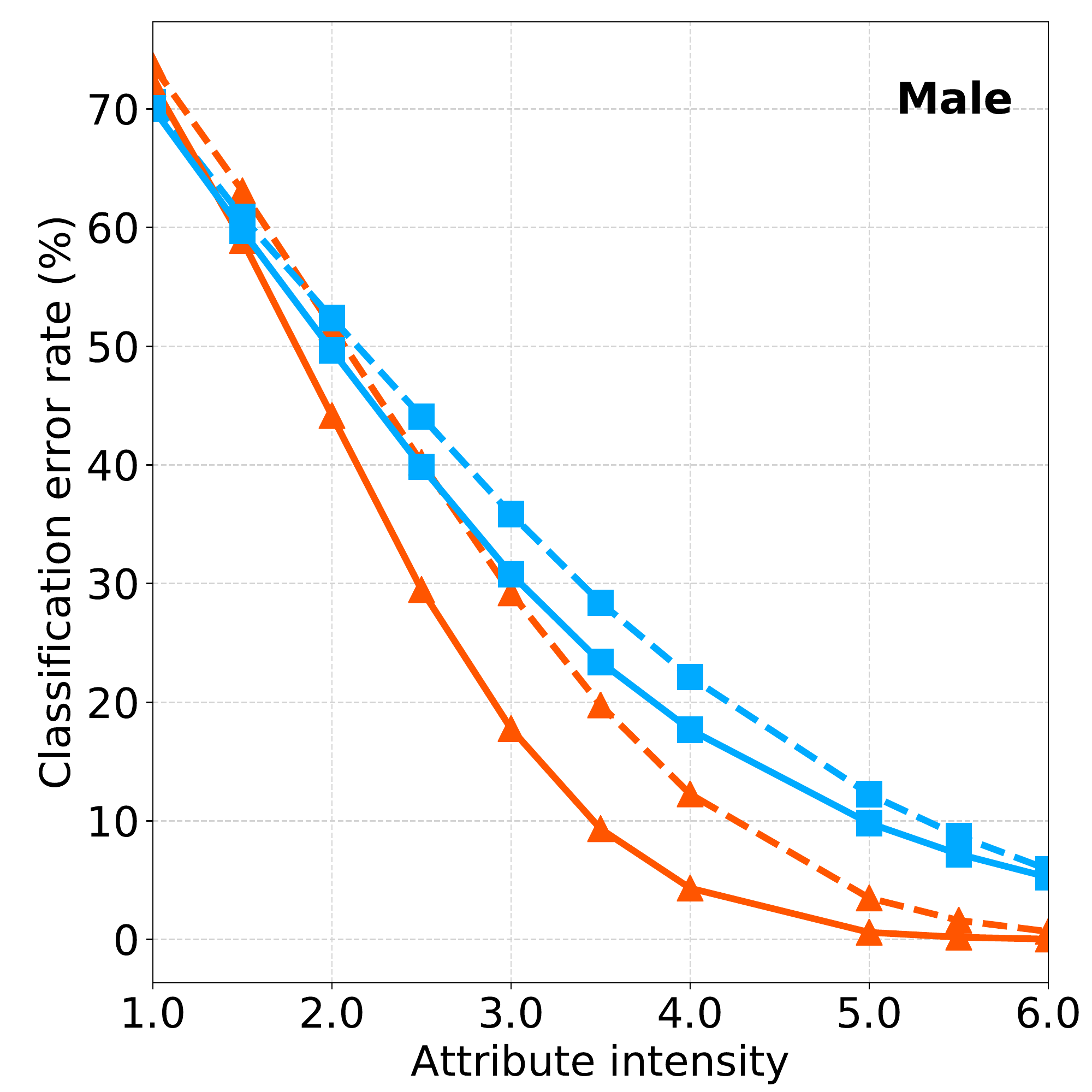}}
	\subfloat[]{\includegraphics[width=0.20\textwidth]{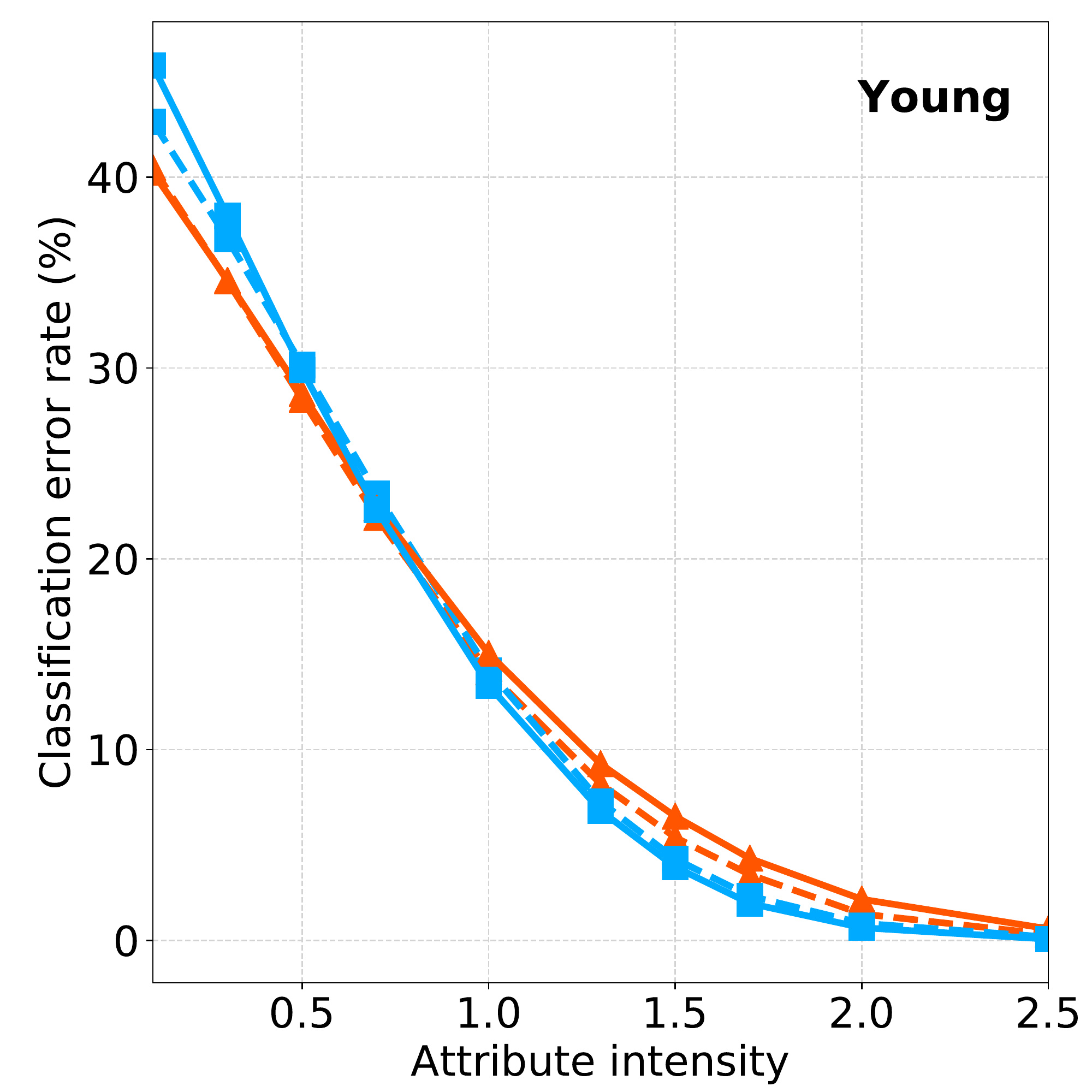}}
	\subfloat[]{\includegraphics[width=0.20\textwidth]{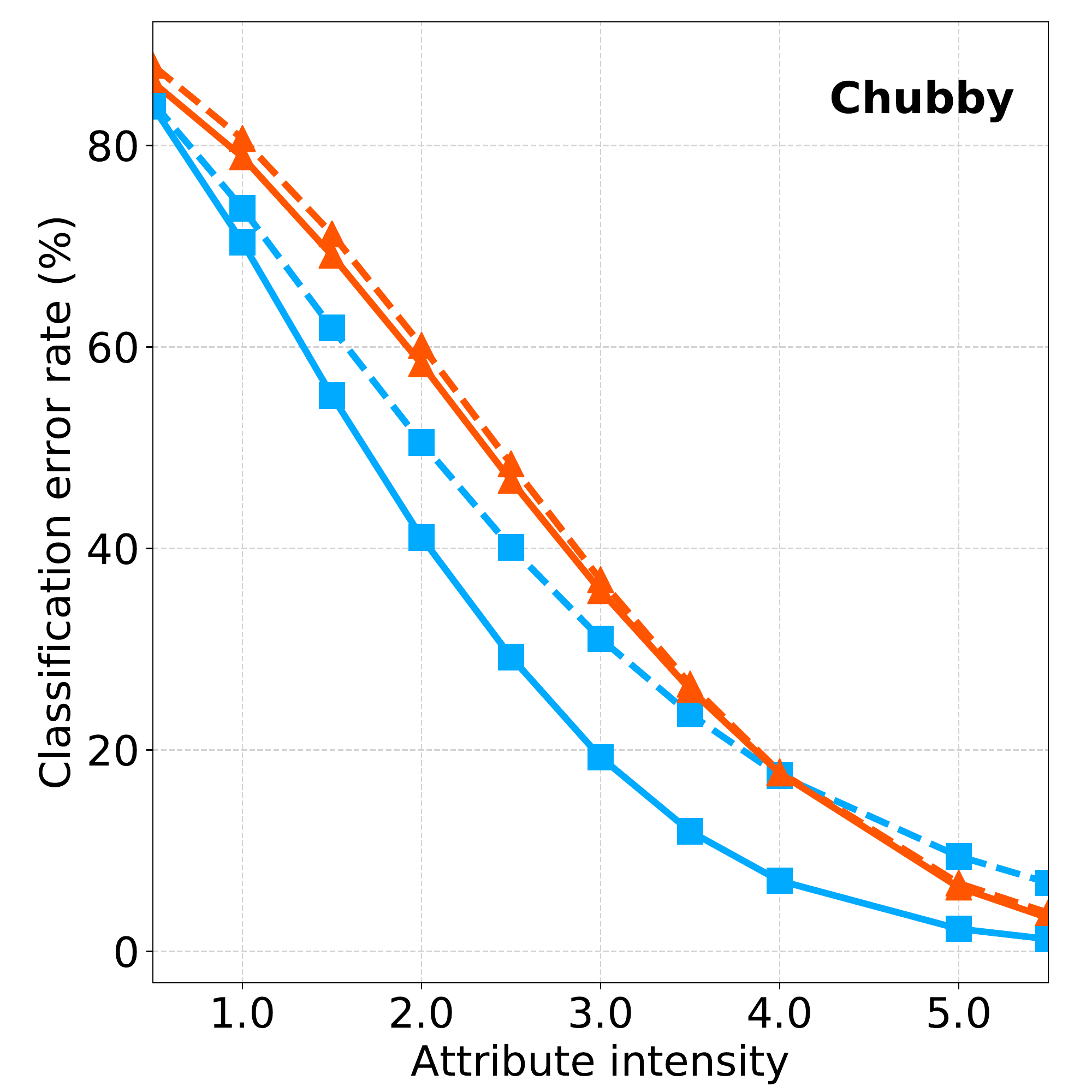}}\\ \vspace{2mm}
	\includegraphics[width=0.8\textwidth]{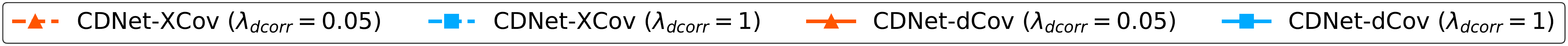}
	\caption{Attribute classification of face images synthesized by CDNet-XCov and CDNet-dCov. (a) Bangs. (b) Receding hairline. (c) Blond hair. (d) Eyeglasses. (e) Bushy eyebrows. (f) Pale skin. (g) Smiling. (h) Mustache. (i) Straight hair. (j) Narrow eyes. (k) Pointy nose. (l) Sideburns. (m) Male. (n) Young. (o) Chubby. For each attribute, we train five linear SVMs independently and report the average performance on each of the ten attribute intensities.}\label{fig:compare_disentangle_strength_by_class}
\end{figure*}

To further analyze the difference between XCov and $\text{dCov}^{2}$ for disentanglement, we design an evaluation protocol to \textit{quantitatively} compare the disentanglement strength of the CDNet-XCov and the CDNet-dCov. We summarize the evaluation procedure into the following four steps.
\begin{enumerate}
	\item We divide the training set into two subsets: the first subset consists of images with the designated attribute, the second one not.
	\item We train a two-class classifier on the two subsets.
	\item We select all test set images that do not contain the designated attribute, then feed them to the disentanglement model to generate their counterparts with the designated attribute and intensity.
	\item Finally, we employ the classifier trained in the second step to classify those images synthesized in the third step, which produces a classification error rate as the evaluation index.
\end{enumerate}

The evaluation protocol is built based on a hypothesis, that is, the classifier is well-trained and therefore lower error rate means it is easier for the classifier to perceive the designated attribute in synthesized images. For each attribute, we train a linear SVM as the two-class classifier to perform attribute classification tasks. Each attribute is assigned ten different intensities, arranged from the lowest level to the highest one, and the two disentanglement models act to synthesize images with both the designated attribute and these corresponding attribute intensities. The main evaluation results are illustrated in Fig. \ref{fig:compare_disentangle_strength_by_class}.

\begin{figure*}[!h]
	\centering
	\subfloat[]{\includegraphics[width=0.38\textwidth]{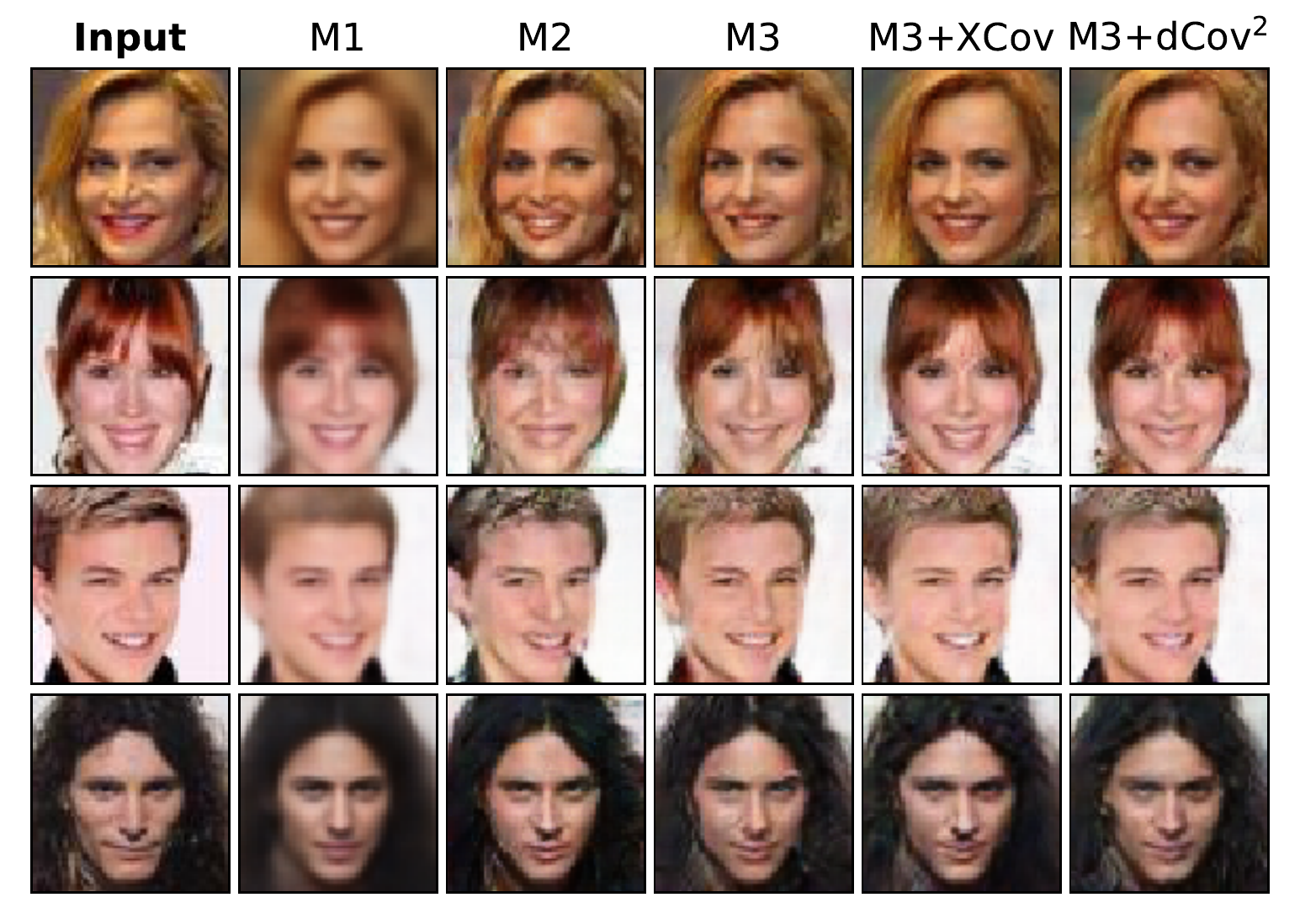}}\hspace{5mm}
	\subfloat[]{\includegraphics[width=0.57\textwidth]{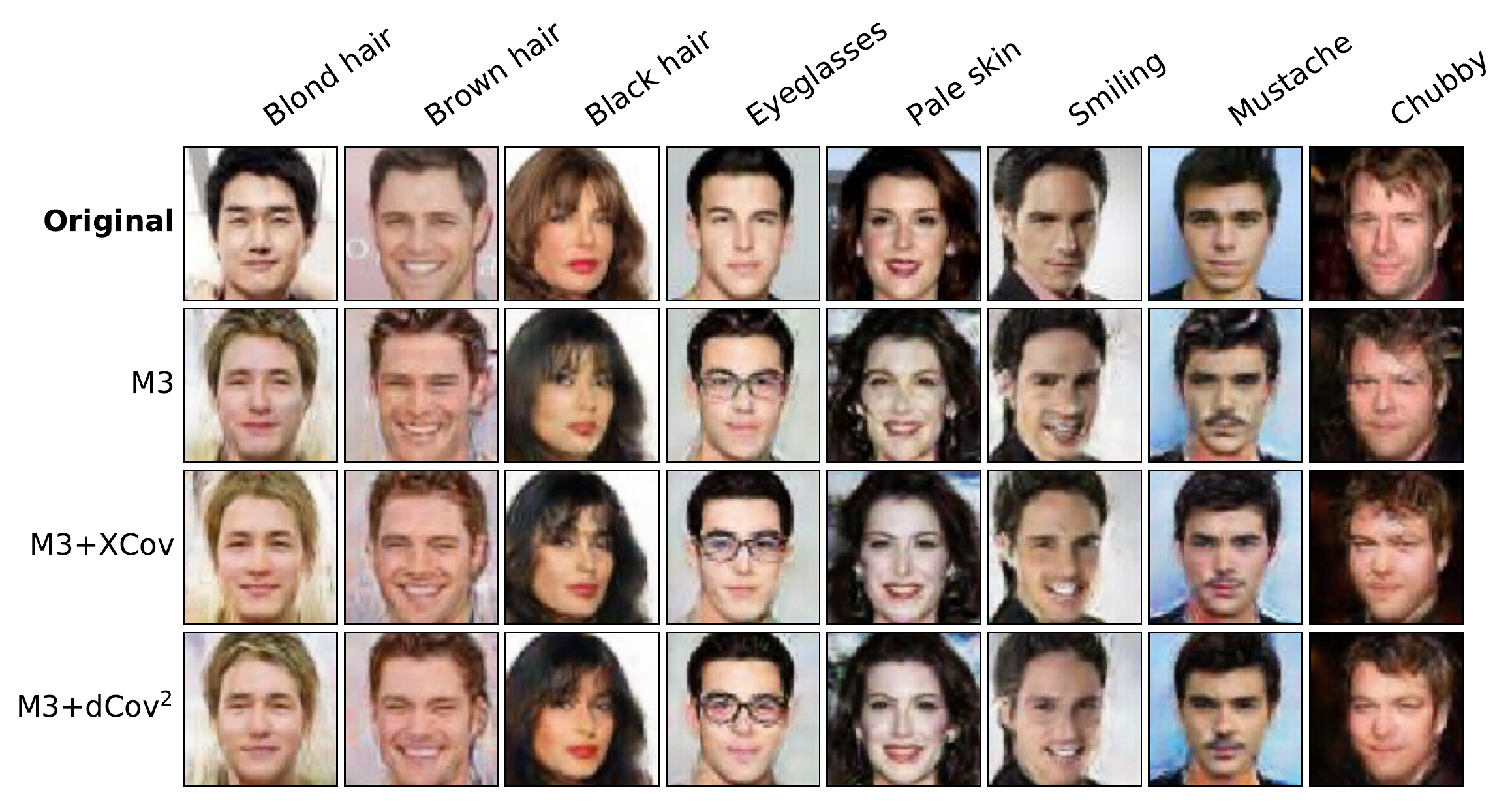}}
	\caption{Qualitative ablation evaluation of different loss terms. (a) Reconstruction results. (b) Disentanglement results. The images are generated by models trained with different local losses. M1: $\mathcal{L}_{class} + \mathcal{L}_{rec}^{pix}$. M2: $\mathcal{L}_{class} + \mathcal{L}_{rec}^{feat} + \mathcal{L}_{adv}$. M3: $\mathcal{L}_{class} + \mathcal{L}_{rec} + \mathcal{L}_{adv}$. Best viewed in color.}\label{fig:ablation_study}
\end{figure*}

As shown in Fig. \ref{fig:compare_disentangle_strength_by_class}, for each facial attribute, the classification error rates consistently decrease as increasing attribute intensities. This result implies that by taking higher attribute intensities, the two CDNets can synthesize images with more distinct attributes. In addition, under the same network architecture and parameter settings, the classification performance of CDNet-dCov is comparable and even superior to that of CDNet-XCov, especially when improving the influence of the decorrelation regularization on learning latent representations (i.e., setting $\lambda_{dcorr}=1$). We attribute this performance gap to the strong disentanglement ability of $\text{dCov}^{2}$, as minimizing $\text{dCov}^{2}$ induces independence between two random variables, rather than the non-correlation as approximated by minimizing XCov. For this reason, in the CDNet-dCov, modifications on the target attributes have less effect on the core identity information contained in the latent representation. This property enables the CDNet-dCov to synthesize images with more concrete and distinguishable attributes, compared with the CDNet-XCov.

\subsection{Ablation Study}\label{sec:ablation_study}
To verify the impact of different loss terms of the proposed model, we conduct an ablation study on the reconstruction task and the disentanglement task, respectively. From the reconstruction results in Fig. \ref{fig:ablation_study}(a), we observe that images generated by the model M1 are quite blurry and only capture the coarse shape of faces; and that images produced by the model M2 show more texture features, but followed with the content deformation in some regions (e.g., hair and mouth). The results demonstrate that minimizing pixel-wise reconstruction loss encourages the model to preserve global identity information, while minimizing feature-wise reconstruction loss is beneficial to restore local detail features. Consequently, models trained with both of these two reconstruction losses perform better in generating more plausible face images, as indicated by reconstruction results of M3, M3$+\text{XCov}$, and M3$+\text{dCov}^{2}$.

For the disentanglement comparison shown in Fig. \ref{fig:ablation_study}(b), the plain model M3 (trained without using any decorrelation term) is limited to synthesize target facial attributes being consistent with the context. For example, changing one attribute, such as ``Smiling'' or ``Mustache'', also causes the considerable image quality degradation near the nose area. Besides, M3 cannot preserve the hairstyle during image editing, which can be found in the synthesized faces corresponding to attributes ``Brown hair'', ``Black hair'', ``Chubby'', and so on. By contrast, when training the same neural network using the additional decorrelation regularization (i.e., M3$+\text{XCov}$ and M3$+\text{dCov}^{2}$), the designated attributes can be blended in new faces properly and with less effect on other attributes. The results suggest that the decorrelation term indeed facilitates models to learn uncorrelated representations during training, thus approaching controllable image editing at test time. Furthermore, we find the synthesized images of M3$+\text{dCov}^{2}$ are more visually realistic and coherent than those of M3$+\text{XCov}$. This difference in performance is observed especially when comparing the hairstyle of subjects under the ``Black hair'' and the ``Chubby'' attributes, or comparing the mouth area of subjects under the ``Brown hair'' attribute. Our conclusion is that compared with the XCov, the $\text{dCov}^{2}$ is more helpful for models to separate the information of interest from other portions, resulting in the ability to synthesize pleasant images in an easier-controllable manner.

\section{Conclusion}
In this paper, we proposed a simple yet effective model that aims to address two disentanglement-related problems: controlling the degree of disentanglement at image editing time, and balancing the disentanglement strength and the reconstruction quality. A distance covariance based decorrelation regularization was devised to encourage disentanglement, and the soft target representation was explored to control how much a specific attribute is perceivable in the generated image. Besides, a method of combining AE with GAN was designed to improve the visual quality of reconstructed and synthetic images. In addition, we also developed a classification protocol to quantitatively evaluate the disentanglement strength of our model. Experimental results demonstrated that our model is able to generate new digits with various handwriting styles, and also to synthesize novel faces with the desired attributes and the attribute intensities.

The supervised disentanglement learning, as well as the decorrelation regularization used in this work, enables the model to learn target representations effectively. However, acquiring a large amount of labeled training data is usually costly and time consuming. To alleviate this problem, we can consider extending the current model to be suitable for the semi-/weakly supervised learning scenario. One possible way to approach this goal is, as discussed in \cite{Kingma14b}, devising probabilistic models for inductive and transductive semi-supervised learning, which can be further implemented by using the approximate Bayesian inference method. Moreover, exploring the concrete benefits of disentangled representations for downstream tasks is another promising direction in this research filed \cite{Locatello19}.

\appendix
\numberwithin{equation}{section}
\setcounter{equation}{0}
We leverage the theoretical results in \cite{Szekely07} to prove Theorem \ref{thm:independ_dcov}. Overall, we use properties of the distance correlation to connect the sample distance covariance (dCov) to the independence between two random variables. The following one definition and two lemmas correspond to the Definition 3, Theorem 2, and Theorem 3 in \cite{Szekely07}, respectively. One can find the complete proofs to the two lemmas therein.
\begin{definition}\label{def:dist_corr}
	The distance correlation between two random vectors $\mathbf{U}$ and $\mathbf{V}$ with finite first moments is the nonnegative number $\mathcal{R}(\mathbf{U}, \mathbf{V})$ defined by
	\begin{equation}
	\mathcal{R}^{2}(\mathbf{U}, \mathbf{V})=\begin{cases}
	\dfrac{\mathcal{V}^{2}(\mathbf{U}, \mathbf{V})}{\sqrt{\mathcal{V}^{2}(\mathbf{U})\mathcal{V}^{2}(\mathbf{V})}}, & \mathcal{V}^{2}(\mathbf{U})\mathcal{V}^{2}(\mathbf{V}) > 0, \\
	0, & \mathcal{V}^{2}(\mathbf{U})\mathcal{V}^{2}(\mathbf{V}) = 0
	\end{cases}
	\end{equation}
	where $\mathcal{V}(\mathbf{U}, \mathbf{V})$ is the distance covariance between $\mathbf{U}$ and $\mathbf{V}$, $\mathcal{V}(\mathbf{U})$ and $\mathcal{V}(\mathbf{V})$ are distance variances of $\mathbf{U}$ and $\mathbf{V}$, respectively.
\end{definition}

\begin{lemma}\label{lemma:limit_dist_cov}
	If $E(\|\mathbf{U}\|_{2}) < \infty$ and $E(\|\mathbf{V}\|_{2}) < \infty$, then almost surely
	\begin{equation}
	\lim_{K \to \infty} \text{dCov}_{K}(\mathbf{U}, \mathbf{V}) \stackrel{a.s.}{=} \mathcal{V}(\mathbf{U}, \mathbf{V}).
	\end{equation}
\end{lemma}

\begin{lemma}\label{lemma:dist_corr_independ}
	If $E(\|\mathbf{U}\|_{2} + \|\mathbf{V}\|_{2}) < \infty$, then $0 \leq \mathcal{R} \leq 1$, and $\mathcal{R}(\mathbf{U}, \mathbf{V}) = 0$ if and only if $\mathbf{U}$ and $\mathbf{V}$ are independent.
\end{lemma}

\begin{IEEEproof}[Proof of Theorem \ref{thm:independ_dcov}]
	According to the given condition that $E(\|\mathbf{U}\|_{2}) < \infty$ and $E(\|\mathbf{V}\|_{2}) < \infty$, Lemma \ref{lemma:limit_dist_cov} holds. Then we have	
	\begin{align}\label{eqn:limit2}
	\lim_{K \to \infty} \text{dCov}^{2}_{K}(\mathbf{U}, \mathbf{V}) &= \left(\lim_{K \to \infty} \text{dCov}_{K}(\mathbf{U}, \mathbf{V})\right)^{2} \nonumber \\ 
	                                                                &\stackrel{a.s.}{=} \mathcal{V}^{2}(\mathbf{U}, \mathbf{V}).
	\end{align}
	The first step in Eq. \eqref{eqn:limit2} is done because the two limitations $\lim_{K \to \infty} \text{dCov}^{2}_{K}(\mathbf{U}, \mathbf{V})$ and $\lim_{K \to \infty} \text{dCov}_{K}(\mathbf{U}, \mathbf{V})$ exist, and the second step simply applies Lemma \ref{lemma:limit_dist_cov}. Notice that we assumed $\lim_{K \to \infty} \text{dCov}^{2}_{K}(\mathbf{U}, \mathbf{V}) = 0$, there holds
	\begin{equation}
	\mathcal{V}^{2}(\mathbf{U}, \mathbf{V}) \stackrel{a.s.}{=} 0.
	\end{equation}
	Based on Definition \ref{def:dist_corr}, we get
	\begin{equation}\label{eqn:dist_corr_zero}
	\mathcal{R}(\mathbf{U}, \mathbf{V}) \stackrel{a.s.}{=} 0.
	\end{equation}
	The condition $E(\|\mathbf{U}\|_{2} + \|\mathbf{V}\|_{2}) < \infty$ in Lemma \ref{lemma:dist_corr_independ} is satisfied as $E(\|\mathbf{U}\|_{2}) < \infty$ and $E(\|\mathbf{V}\|_{2}) < \infty$. Thus, from Lemma \ref{lemma:dist_corr_independ} and Eq. \eqref{eqn:dist_corr_zero}, almost surely $\mathbf{U}$ and $\mathbf{V}$ are independent. This completes the proof of Theorem \ref{thm:independ_dcov}.
\end{IEEEproof}

\section*{Acknowledgment}
The authors would like to thank the anonymous reviewers for their valuable comments and suggestions. They also thank Microsoft Azure for computing resources.

\ifCLASSOPTIONcaptionsoff
\newpage
\fi

\bibliographystyle{IEEEtran}
\bibliography{references}

\end{document}